\documentclass[11pt]{article}

\usepackage[preprint]{acl}

\usepackage{times}
\usepackage{latexsym}
\usepackage{tikz}
\usetikzlibrary{arrows.meta, positioning}
\usepackage{booktabs} 
\usepackage{multirow} 
\usepackage{xcolor}
\usepackage{float}
\usepackage{soul} 
\usepackage{subcaption}
\usepackage{todonotes}
\usepackage[T2A,T1]{fontenc}
\usepackage[utf8]{inputenc}
\usepackage[russian,english]{babel}

\usepackage{microtype}

\usepackage{inconsolata}
\usepackage{color-edits}
\addauthor[Alexis]{ap}{cyan}
\addauthor[Ali]{am}{teal}
\addauthor[Katharina]{kw}{blue}

\usepackage{tikz}

\usepackage[normalem]{ulem}
\usepackage{graphicx}

\definecolor{lightred}{RGB}{255, 120, 120}
\definecolor{darkred}{RGB}{150, 0, 0}

\newcommand{\err}[2]{%
  \tikz[baseline=(char.base)]{
    \node[rectangle, fill=#2!40, thick, inner sep=2pt] (char) {#1};
  }
}

\title{Speaking in Self-Assessing Tongues: On the Verbalized Confidence of LLMs in Machine Translation}

\author{
 \textbf{Ali Marashian\textsuperscript{1}} \qquad
 \textbf{Alexis Palmer\textsuperscript{1}} \qquad
 \textbf{Katharina von der Wense\textsuperscript{1, 2}} 
\\
 \textsuperscript{1}University of Colorado Boulder, USA \\
 \textsuperscript{2}Johannes Gutenberg University Mainz, Germany
\\
 \small{
   \textbf{Correspondence:} \href{mailto:ali.marashian@colorado.edu}{ali.marashian@colorado.edu}
 }
}

\begin{document}
\maketitle
\begin{abstract}
The rapid rise in popularity of large language models (LLMs) for translation calls for a thorough study of the reliability of their confidence in their own outputs.  
Unlike many generation tasks, translation errors and confidence levels can be useful at different levels of granularity (tokens, words, or spans). Unsupervised approaches based on internal signals like predicted probabilities can be misleading because they reflect certainty among alternatives rather than correctness. In addition, they require access to such internal signals.  Here, we devise five \textit{verbalized} methods of extracting an LLM's per-token confidence without those shortcomings and compare their reliability with that of the model's internal signals of certainty. 
We evaluate reliability using two forms of alignment: fine-grained error detection and calibration. For both, internal and verbalized methods perform similarly, although results vary by model.
Interestingly, we find little to no correlation between internal and verbalized methods.

\end{abstract}

\section{Introduction}

With the growing use of large language models (LLMs) for machine translation (MT)  \cite{kocmi-etal-2024-findings, kocmi-etal-2025-findings}, it is 
becoming increasingly important to be able to judge their reliability. However, LLMs are known to be poorly self-calibrated for MT: output token probabilities are not good indicators of the quality of the corresponding translation \cite{wu2025calibrating, sarti-etal-2025-unsupervised}. 

\begin{figure}[h]
\begin{center}
\includegraphics[width=1\linewidth]{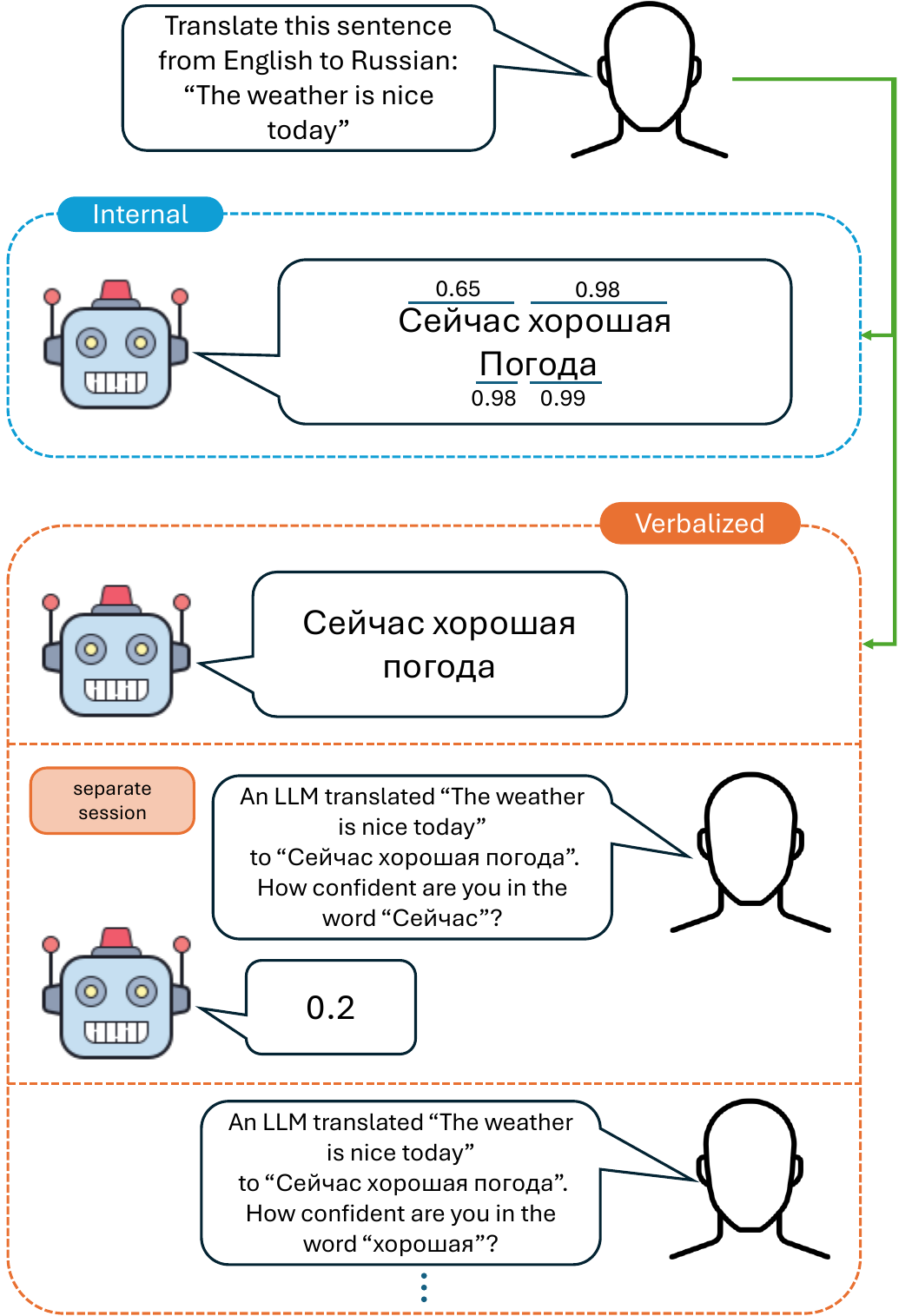}
\end{center}
\caption[align=center]{
Illustration of internal vs. verbalized confidence estimation, using token probabilities as internal signals and word-level confidence obtained via prompts; we explore broader variants of both in our experiments.
}
\label{fig:overall_verbalized}
\end{figure}

%

One of the key challenging aspects of MT calibration is that there is not just one single way of translating any given input -- e.g., one can choose between synonyms for  translation of the same concept, or between multiple syntactic realizations of the same semantic content. Therefore, 
if the model is unsure between multiple possible tokens, we cannot directly consider that  as representing the model's confidence in the validity of the selected token. Instead, it  demonstrates the model's \textit{uncertainty}, as some potentially valid options all compete with respect to probability. More generally, this phenomenon in generative tasks is known as \textit{surface form competition} \cite{holtzman-etal-2021-surface, wiegreffe-etal-2023-increasing}. 

Something which so far is underexplored in MT is that LLMs have the unique ability to \textit{verbalize} how certain they are by generating tokens that estimate confidence \cite{tian-etal-2023-just} -- i.e., by "saying" how confident they are. Thus, 
they can provide an estimate of the correctness of their outputs that is not limited by surface form competition. Verbalized confidence has the further advantage of being easily obtainable by end users and without access to model internals.


In this paper, we focus on the verbalized per-token confidence of LLMs for MT and explore how it aligns with the LLMs' translation performance (cf. Figure~\ref{fig:overall_verbalized}). 
Since there is no single, programmatic way of accessing verbalized confidence, we explore five techniques. 
For (the commonly considered) internal uncertainty, we employ two different indicators: token-level probability and entropy. We compare how both verbalized confidence and internal uncertainty align with the ground truth using two evaluation paradigms: fine-grained binary error detection and calibration. Additionally, we examine the correlation between the verbalized confidence and internal uncertainty to understand how similar the two high-level strategies are. 

Our results show that verbalized confidence performs comparably to or better than internal uncertainty-based methods on fine-grained binary error detection on average, depending on the model. For calibration, we find that verbalized methods perform similarly to model probability but lag behind model entropy.  
Moreover, we find no significant correlation between verbalized confidence and internal uncertainty in our setting.

\section{Related Work}
\label{related_work}
Although \textit{uncertainty} (the variability of a model's response for a particular input) and \textit{confidence} (the model's belief that a specific output is correct) are distinct concepts, they are so closely related that uncertainty is frequently used in practice to estimate confidence \cite{liu2025uncertainty}. As a result, ``calibration'' in the literature can mean both the alignment between confidence (e.g., as verbalized by the model itself) and ground-truth and the alignment between certainty and ground-truth. 

To estimate model uncertainty, some approaches utilize external knowledge, e.g., auxiliary models trained to predict confidence scores \cite{mielke-etal-2022-reducing, tsai2024efficient, ulmer-etal-2024-calibrating}, or using search engines to detect errors \cite{gou2023critic}. 
Other approaches rely solely on the model. A number of them quantify sample consistency, measuring the similarity between multiple model responses 
\cite{tian-etal-2023-just, manakul-etal-2023-selfcheckgpt, wang2024subjective}. 
Many works use internal signals. For multiple-choice question answering or other classification tasks, one can aggregate the probabilities of different tokens that the model would use for each option \cite{han2022prototypical, zhang-etal-2024-calibrating, wang2024calibrating, lovering2024language, kumar-etal-2024-confidence}. 


Another approach that leverages only the model itself is to have the LLMs verbalize their confidence. This is typically done for QA
\cite{lin2022teaching, kadavath2022language, tian-etal-2023-just, yang2024verbalized, xiong2023can, ni2024large}. Instead of QA, our work focuses on MT. Rather than considering the ground truth at all,
 \citet{kumar-etal-2024-confidence} focus on the relationship between model's internal uncertainty and its verbalized confidence for QA, introducing the concept of ``Confidence-Probability Alignment.'' In our work, we also consider the correlation of verbalized methods with token \textit{entropies} as an additional internal signal.

MT calibration has been a topic of interest in the literature even before the boom of LLMs \cite{ott2018analyzing, kumar2019calibration, wang-etal-2020-inference, lu-etal-2022-learning}. A key challenge for MT calibration is assigning errors to particular spans in the output. Previous work has solved the problem by automatically estimating these spans using Translation Edit Rate \cite{snover-etal-2006-study} \footnote{By  considering the minimum number of edits necessary to get from the produced hypothesis to a gold translation: if a token needs to change, it is  labeled as wrong. These pseudo-labels have been shown to be inconsistent with real labels \cite{yang-etal-2023-rethinking}.} or manually annotating the output of MT models \cite{fomicheva-etal-2022-mlqe, sarti-etal-2022-divemt, yang-etal-2023-rethinking, sarti2025qe4pe}. 
\citet{dinh-niehues-2025-generative} boost the probability of certain tokens to counterbalance the resultant under-confidence caused by surface form competition for MT and other generative tasks.


The most relevant study to ours is \citet{sarti-etal-2025-unsupervised}, 
which studies the predictive power of different signals of model uncertainty and their alignment with actual error spans in translation. 
We extend their work by examining the capabilities of LLMs to verbalize their confidence and the alignment of these verbalizations with both ground-truth error spans and internal indicators. 
 
\section{Experimental Design}

We now describe the experimental design, including the datasets, models, different measures of model confidence, and evaluation metrics.


\subsection{Dataset} 

Our goal is to investigate the alignment of different measures of model confidence  with  ground truth error annotations. 
The basis of our experiments is the WMT2024 General Machine Translation shared task data. We look at translations from English to Chinese, Czech, Hindi, Japanese, and Russian. 

In order to assess the reliability of model confidence measures, we further leverage the Error Span Annotations \citep[ESA;][]{kocmi-etal-2024-error}
by \citet{kocmi-etal-2024-findings},
who mark errors in the outputs of several models for 9 language pairs. In ESA, annotators mark erroneous spans in the translation and categorize them as either \textit{minor} or \textit{major} errors. Table \ref{tab:dataset_example} shows an annotated example.

Following prior work, we do not distinguish between the two error categories:
tokens are considered either correct or incorrect \cite{sarti-etal-2025-unsupervised}. For each translation direction--model pair, we have 634 annotated outputs. We use these annotations as the ground truth. As a development set, we randomly select 100 English sentences and their respective translations for each language pair. 
This leaves us with 534 test sentences per language pair. 

\paragraph{Word Segmentation} Some of our proposed verbalized approaches require prompting a model for its confidence for specific words. For this, we automatically segment the text of the translations into words: for Czech, Hindi, and Russian we split text on whitespaces, for Chinese we use Jieba,\footnote{\url{https://github.com/fxsjy/jieba}} and for Japenese we employ Nagisa.\footnote{\url{https://github.com/taishi-i/nagisa}}

\subsection{Models} 
Using ESA, \citet{kocmi-etal-2024-findings} annotate the outputs of 28 systems; including shared task submissions, 7 popular LLMs, and additional online systems. Note that annotations are not available for all systems and all translation directions. Of the available LLMs, we select the only two that are open-source: Aya23 \cite{aryabumi2024aya} and Llama3-70B \cite{grattafiori2024llama}. We have the same setup and prompts as \citet{kocmi-etal-2024-findings}, in order to obtain identical probability and entropy for the annotated outputs.
Code and results are available on GitHub.\footnote{Link will be provided upon publication.} 



\subsection{Verbalized Confidence Measures}
\label{verb:conf}
We now describe our five proposed approaches for obtaining a model's verbalized confidence in its translations.  Following \citet{kumar-etal-2024-confidence}, we use separate prompts for translation and confidence elicitation: we do not provide the information to the LLM that it is evaluating its own translations, in order to avoid bias in its judgments \cite{zheng2023judging}. We leave same-session self-evaluation to future work. 

\begin{enumerate}
  \item \textbf{List}: We prompt the model to return a list containing those spans in the translation that it is not confident about. If a token is part of any returned span, we assign that token the label \textit{incorrect}. 
  
  \item \textbf{Word\_Numeric}: We prompt the model to return its confidence for every word as a number within $[0, 1]$. 
  We ask for the score of each word in a separate prompt; see Figure \ref{fig:overall_verbalized}. For our analysis, the numerical confidence score assigned to a word will be assigned to all of its constituent tokens. If a token contains letters of two words, we treat it as part of the second word. 

  \item \textbf{Word\_Likert}: Instead of numbers, we ask LLMs to assign qualitative labels to each word. Following \citet{kumar-etal-2024-confidence}'s approach for question answering, we consider six different labels: \textit{very uncertain}, \textit{not certain}, \textit{somewhat certain}, \textit{moderately certain}, \textit{fairly certain}, and \textit{very certain}. We convert these labels into real numbers from 0 (for \textit{very uncertain}) to 1 (\textit{very certain}) in increments of 0.2. As in Word\_Numeric, the score assigned to a word is assigned to its constituent tokens. \citet{kumar-etal-2024-confidence}'s motivation for this approach is the hypothesis that an LLM might use qualitative descriptors better than numerical values.

  \item \textbf{Token\_Numeric}: 
  We further explore the effect of increased granularity by using the model's tokenizer and repeating Word\_Numeric with tokens instead of words.\footnote{This requires using the model's tokenizer, which limits its applicability in practical settings. In addition it is more computationally expensive than the other methods. We report the costs of all prompts in Appendix \ref{prompt_cost_report}.}

  \item \textbf{Token\_Likert}: We further experiment with a version of Word\_Likert, but for the outputs provided by the model's tokenizer.

\end{enumerate}

\paragraph{Prompts} We use the development set of English to Czech translations of Aya23 together with the corresponding error annotations for prompt engineering. Appendix \ref{sec:appenidx_prompts} showcases the prompts for all verbalized confidence methods.

\paragraph{Binarization} Since the ground truth labels for error detection are binary, we binarize the output of continuous methods to measure their alignment with the ground truth. For each method--model pair, following \citet{ni2024large}, we tune a threshold on the development set.\footnote{We also report results using an optimal binarization threshold based on the test set itself, rather than using the threshold of the development as we do here. See Appendix \ref{extra_results}.} 

\begin{table}
    \centering
    \small
    \begin{tabular}{lp{5cm}}
        \toprule
        \textbf{Source} & And it turns out that level was uploaded with a TAS after all, so it is no longer considered to count and they've declared victory. Kinda anticlimactic tbh but GG to the people who got everything to that point! \\
        \midrule
        \textbf{Aya23} & \foreignlanguage{russian}{Оказывается, этот уровень \err{был пройден с использованием }{lightred}
        \err{TAS}{lightred}, поэтому он больше не считается, и они объявили победу. Честно говоря, довольно \err{анемично}{darkred}, но \err{ББ}{darkred} тем, кто добрался до этой точки!} \\
        \midrule
        \textbf{Llama3-70B} & \foreignlanguage{russian}{И оказалось, что уровень был загружен с помощью TAS \err{после всего}{darkred}, поэтому он больше \err{не считается засчитанным}{lightred}, и они объявили о победе. Довольно \err{антиклиматично}{darkred}, если честно, но \err{GG}{darkred} людям, которые довели дело до этого момента!} \\
        \bottomrule
    \end{tabular}
    \caption{An example of English $\rightarrow$ Russian translations from WMT24 dataset \cite{kocmi-etal-2024-findings} annotated according to ESA protocol. Light red indicates minor errors, whereas darker red denotes major errors. All tokens that are part of a shaded word are considered wrong in our analysis.}
    \label{tab:dataset_example}
\end{table}

\subsection{Baseline Confidence Measures}

\paragraph{Internal (Un)certainty}
We compare our verbalized approaches to two internal confidence measures. The first one is the \bl{probability} of $t^*_i$,  the 
 $i$-th predicted token in the output translation, given all previously generated tokens: $p(t^*_i|t^*_{<i})$.
The second is the \bl{entropy} of the probability distribution over the vocabulary $V$: $-\sum_{j=1}^{|V|}p(t_{ij}|t^*_{<i}) \, \log_2 p(t_{ij}|t^*_{<i})$.  
A higher probability indicates greater certainty in the correctness of that token.  
In contrast, higher entropy corresponds to lower certainty.  We perform the binarization as outlined in \ref{verb:conf}.

We choose those two measures as they are the top-performing unsupervised methods for the dataset we experiment with \cite{sarti-etal-2025-unsupervised}. In addition to their high performance, they are often obtainable even from closed-source models as those frequently return token probabilities as well as the top-$k$ tokens, from which entropy can be estimated \cite{manakul-etal-2023-selfcheckgpt}.



\paragraph{Random Baseline} 
We further compare to a random baseline that assigns a label to each token by sampling from the labels' probability distribution estimated on the development set.
We repeat this baseline 10 times and report the average.


\subsection{Binary Error Detection}
\label{def:alignment}

To examine how well different confidence measures align with the ground truth, we consider a binary error detection task and report both performance as well as model calibration. 
The goal of this task is to identify \textit{correct} and \textit{incorrect} tokens in a provided translation. 

Following \citet{sarti-etal-2025-unsupervised}, we report the \bl{F1} score of the \textit{incorrect} class as our main metric. We use the binarized version of all confidence metrics. 



We also assess how well the models are calibrated for different confidence measures: in a well-calibrated model, token confidence values accurately correspond to the actual probability of those tokens being correct. 
To measure calibration, we report \bl{Expected Calibration Error (ECE)}. For ECE, the predictions are sorted and partitioned into $K$ bins $\{ B_1, B_2, \cdots, B_K\}$. Each bin corresponds to an interval of confidence. ECE is the weighted average of the absolute difference between the actual accuracy and the expectation of accuracy (according to confidence scores) of all bins:

\[
ECE = \sum_{i=1}^{K} \frac{|B_i|}{N} |acc_{i} - conf_{i}|,
\]

where $B_i$ denotes the number of predictions in bin $i$, $N$ is the total number of predictions, $acc_i$ is the true fraction of correctly predicted instances in bin $i$, and $conf_i$ is the mean of the prediction confidences in bin $i$ \cite{naeini2015obtaining}.

In addition, we report \bl{Area Under the Receiver Operating Characteristic Curve (AUROC)} and \bl{Area Under the Precision-Recall Curve (AUPRC)} scores. They both aggregate the results over all possible thresholds for classification, reflecting how well the confidence scores discriminate between classes \cite{ulmer-etal-2024-calibrating}. AUPRC is more robust to class-imbalance than AUROC \cite{qi2021stochastic}.



\begin{figure}[t]
    \centering
    \begin{subfigure}[b]{0.23\textwidth}
        \centering
        \includegraphics[width=\textwidth]{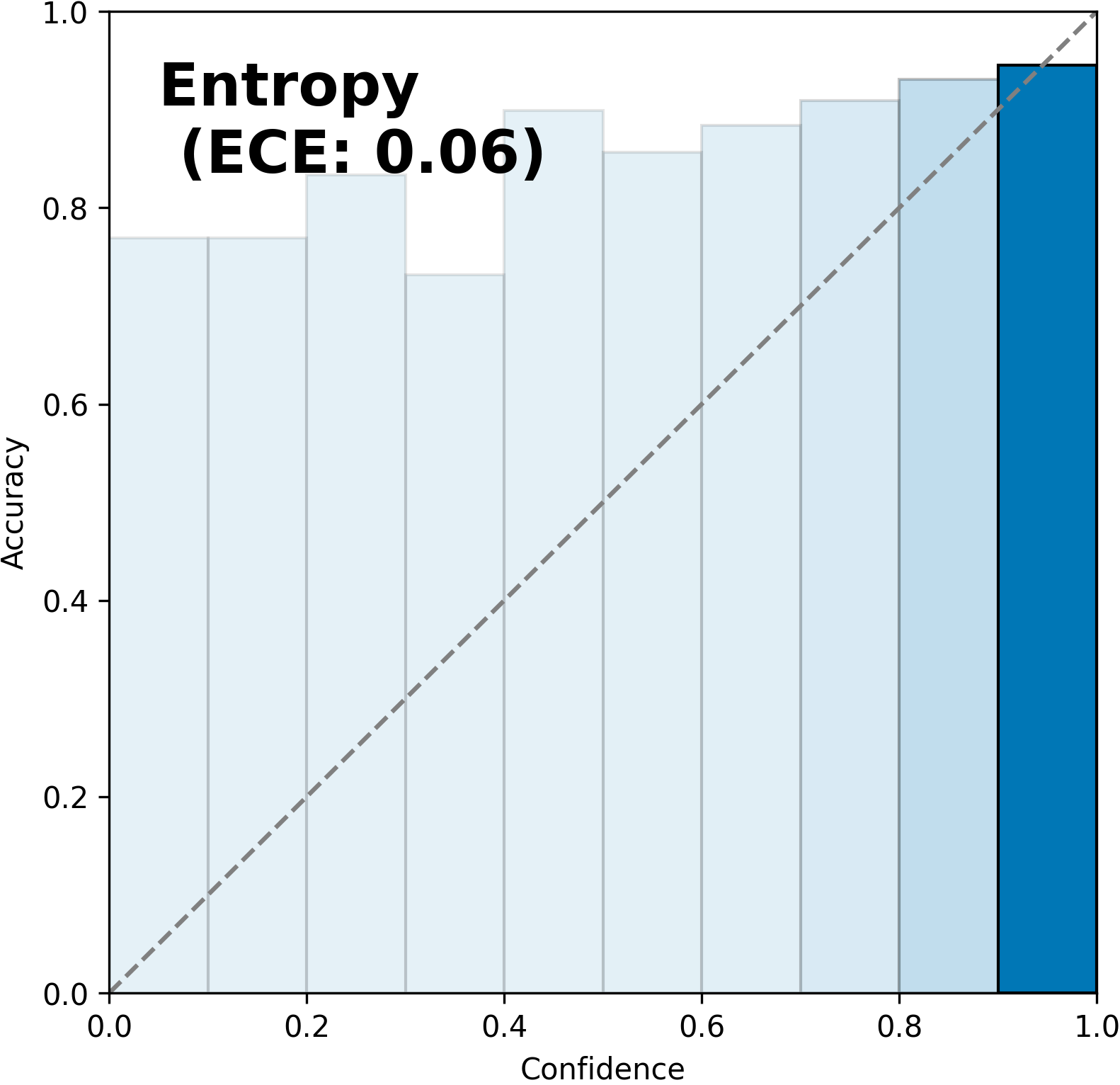}
    \end{subfigure}
    \hfill
    \begin{subfigure}[b]{0.23\textwidth}
        \centering
        \includegraphics[width=\textwidth]{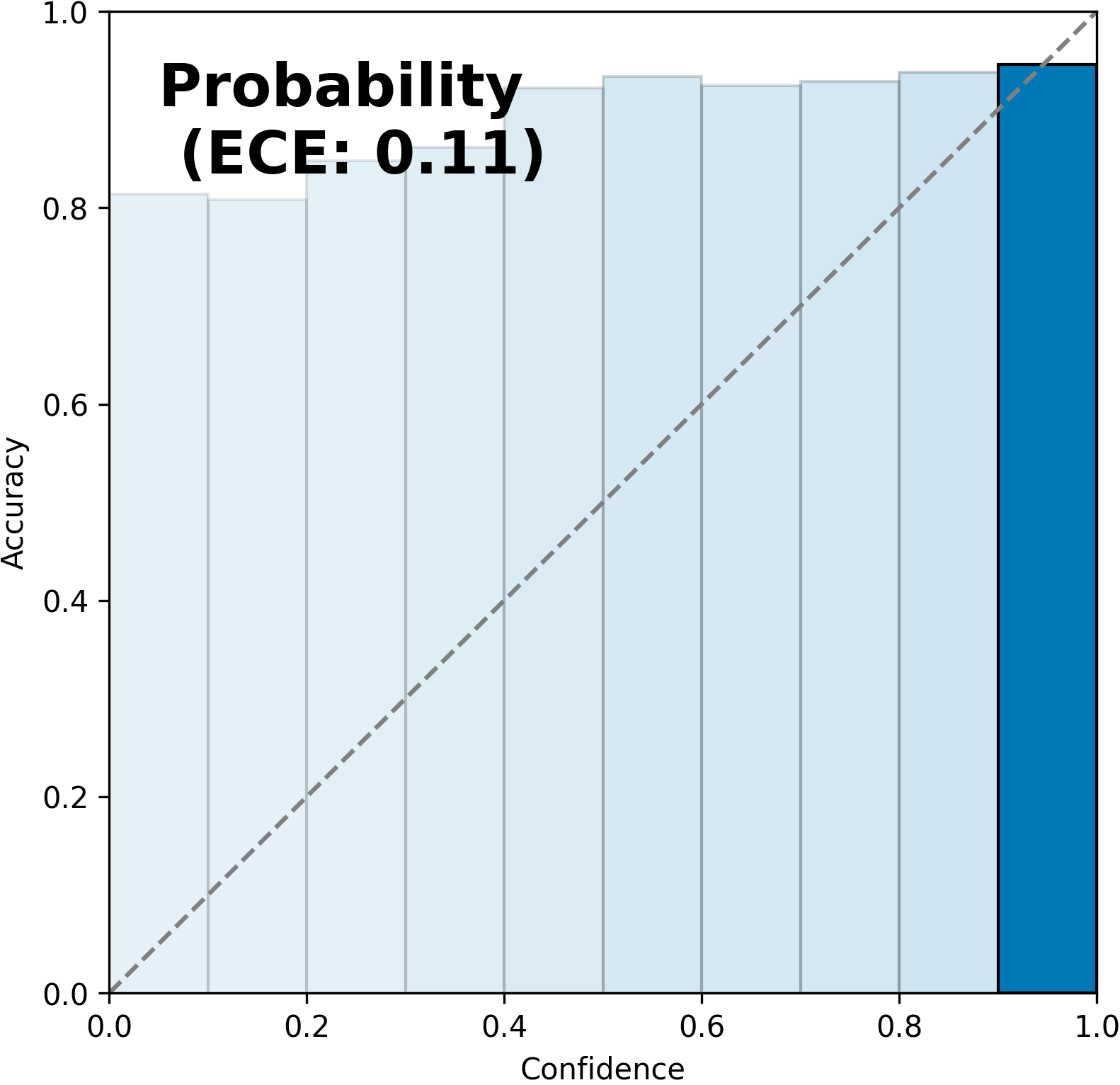}
    \end{subfigure}
    
    \begin{subfigure}[b]{0.23\textwidth}
        \centering
        \includegraphics[width=\textwidth]{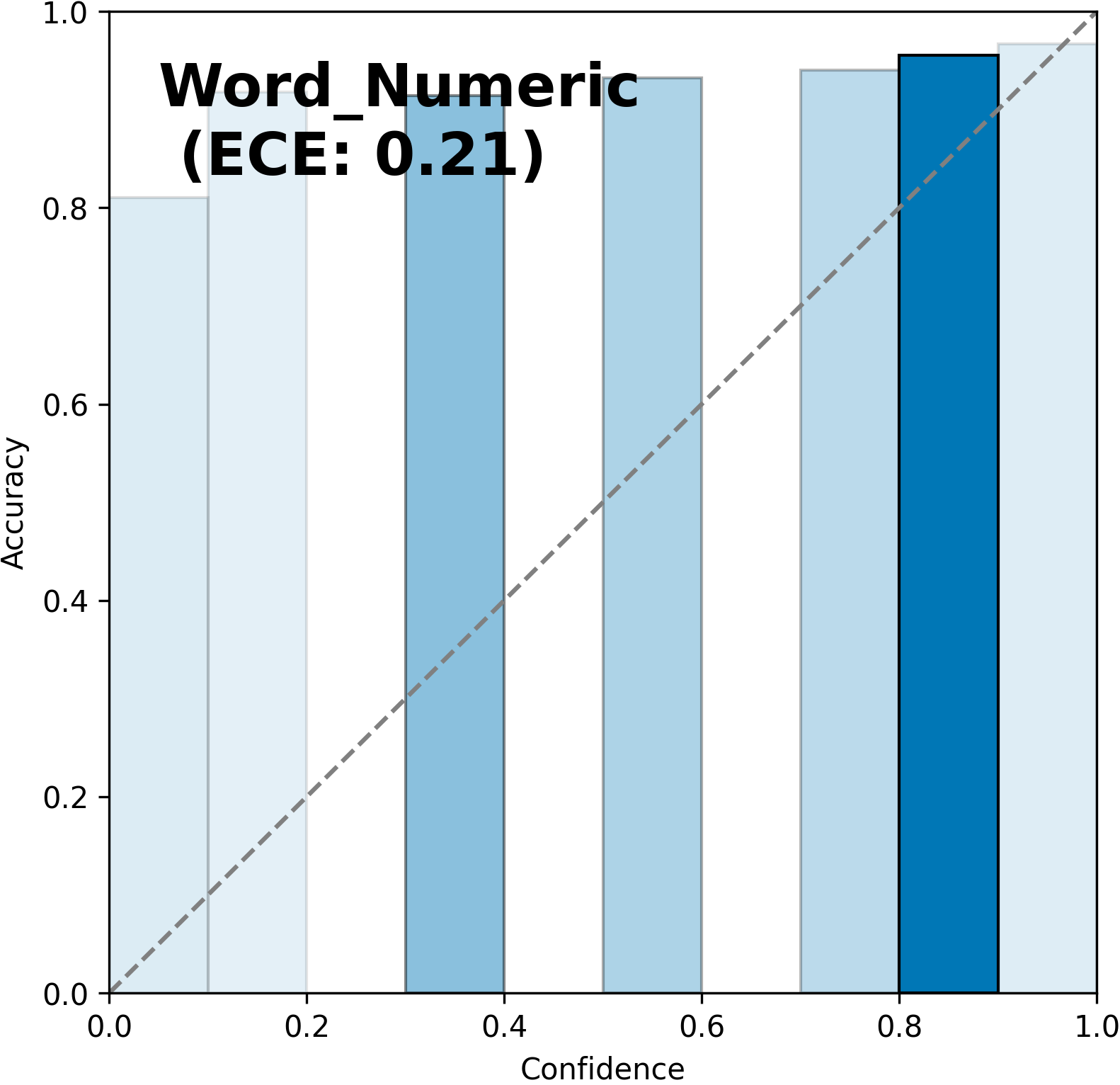}
    \end{subfigure}
    \hfill
    \begin{subfigure}[b]{0.23\textwidth}
        \centering
        \includegraphics[width=\textwidth]{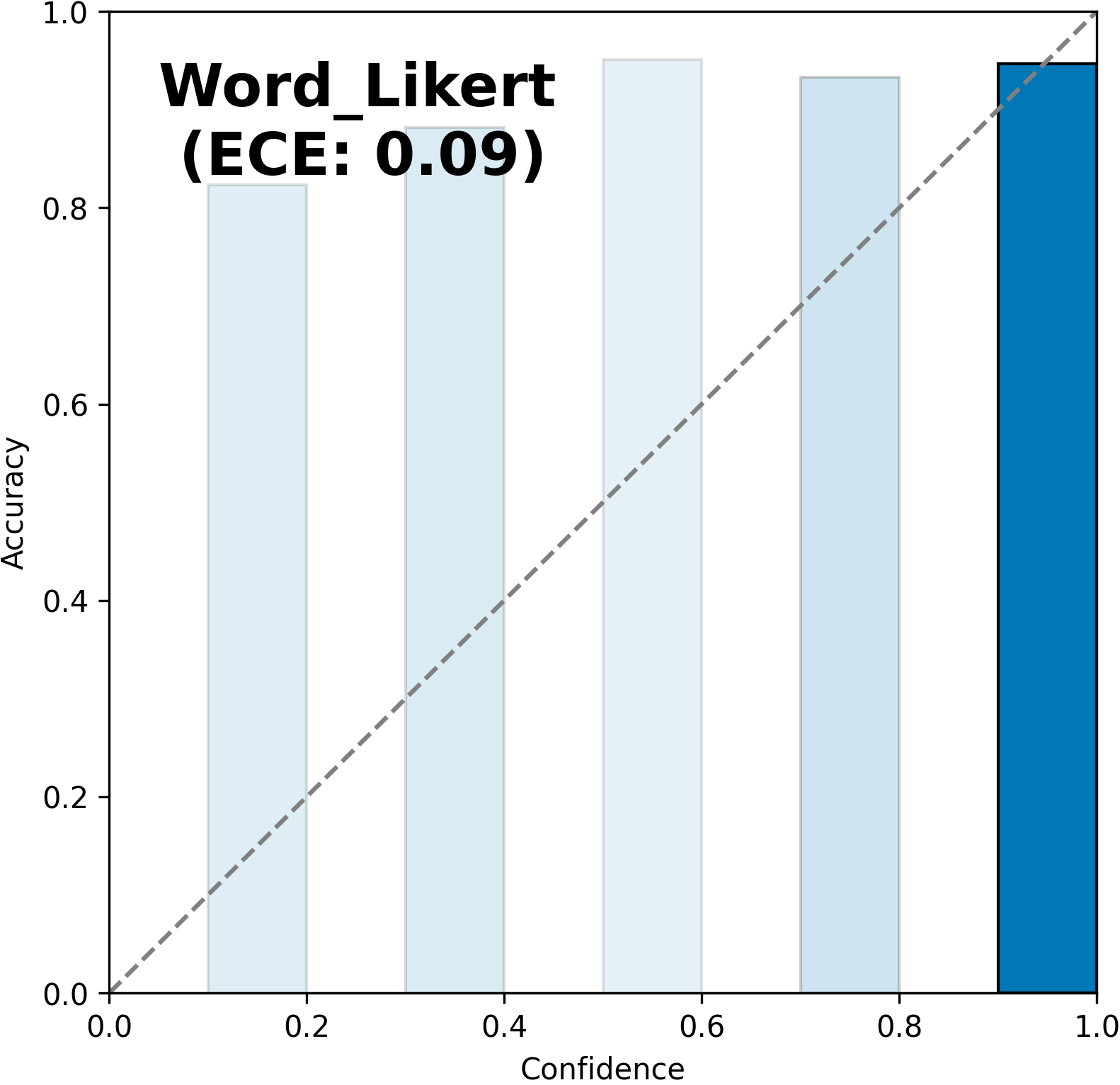}
    \end{subfigure}

    \begin{subfigure}[b]{0.23\textwidth}
        \centering
        \includegraphics[width=\textwidth]{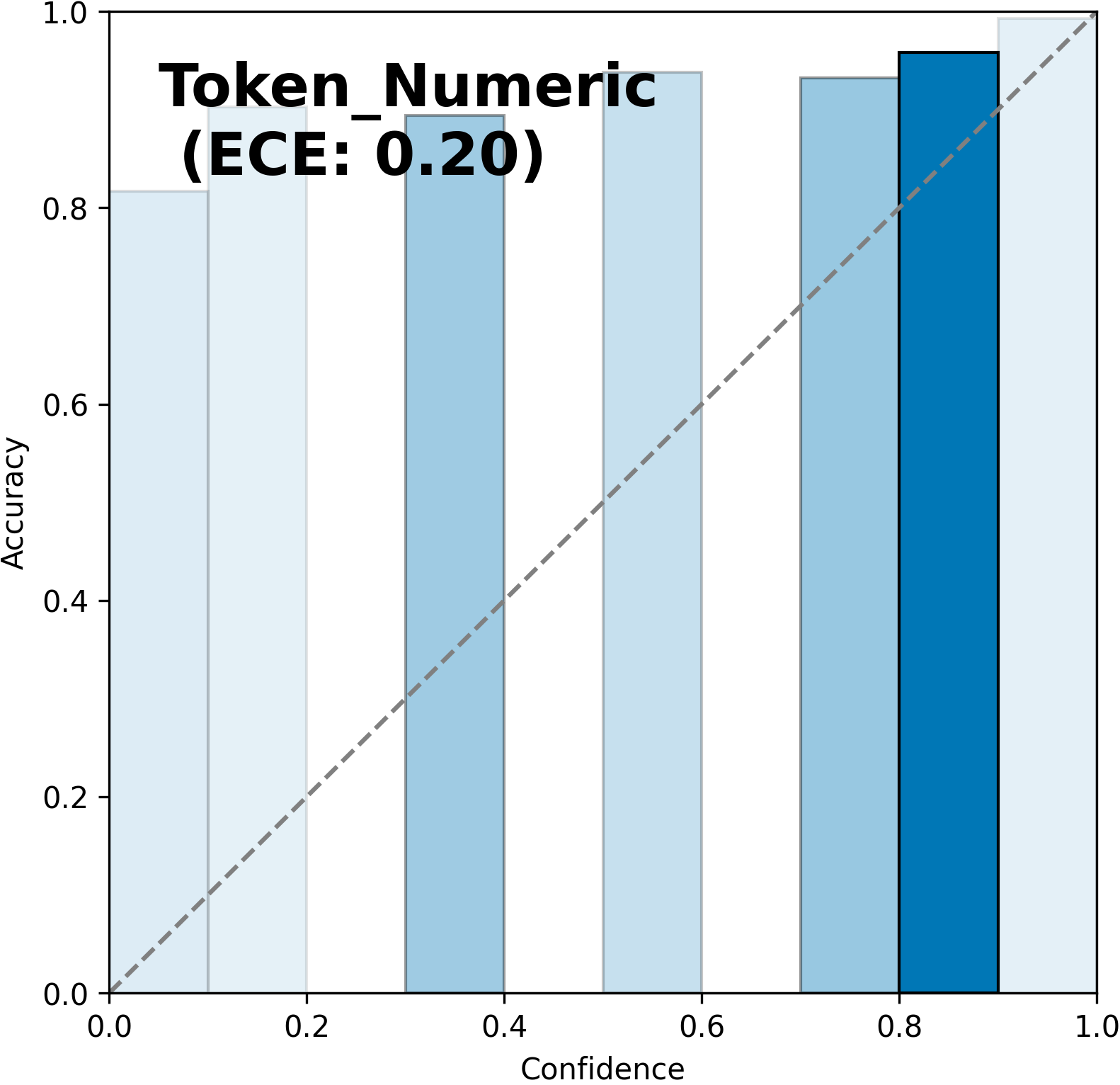}
    \end{subfigure}
    \hfill
    \begin{subfigure}[b]{0.23\textwidth}
        \centering
        \includegraphics[width=\textwidth]{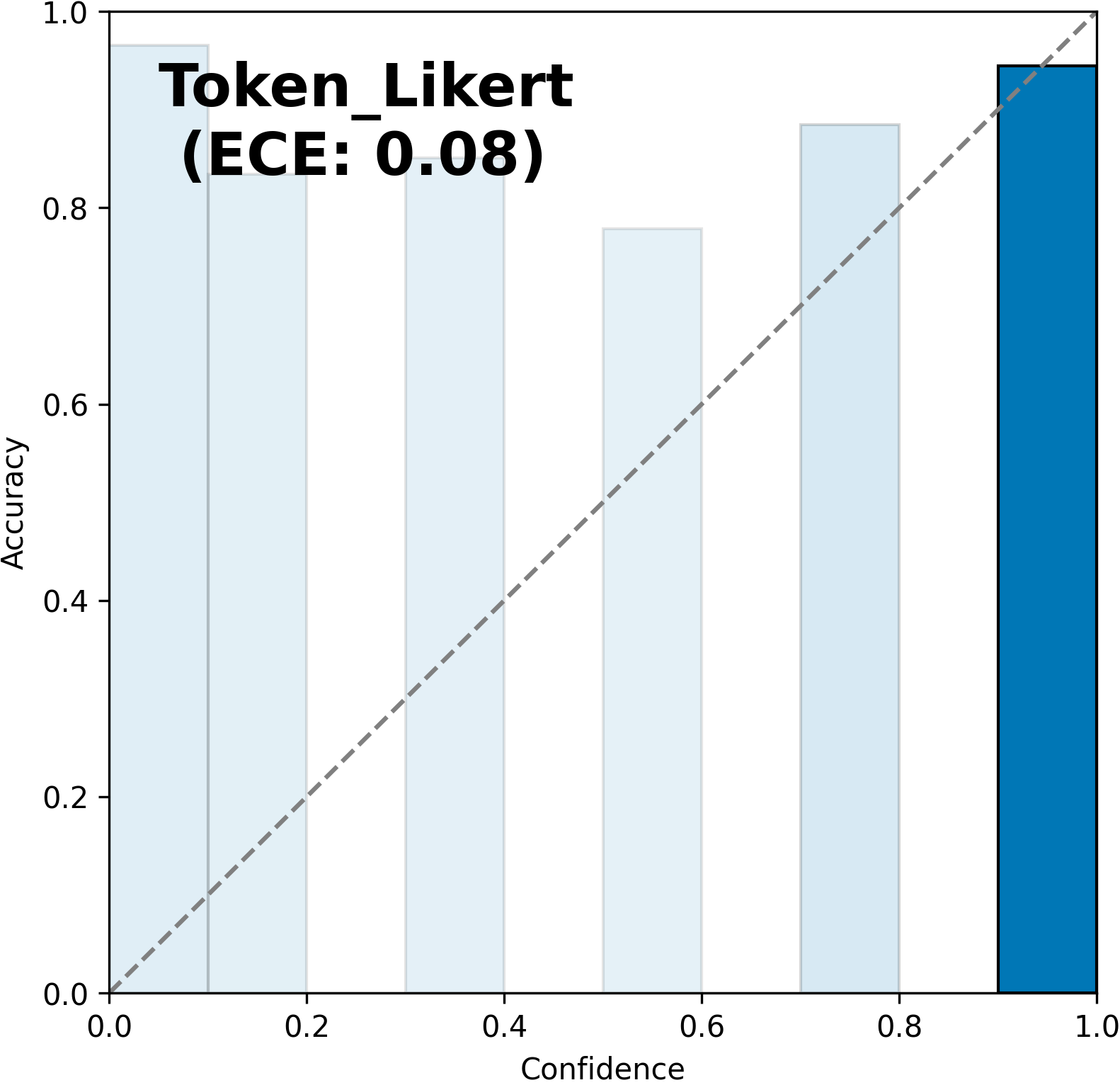}
    \end{subfigure}

    \begin{subfigure}[b]{0.23\textwidth}
        \centering
        \includegraphics[width=\textwidth]{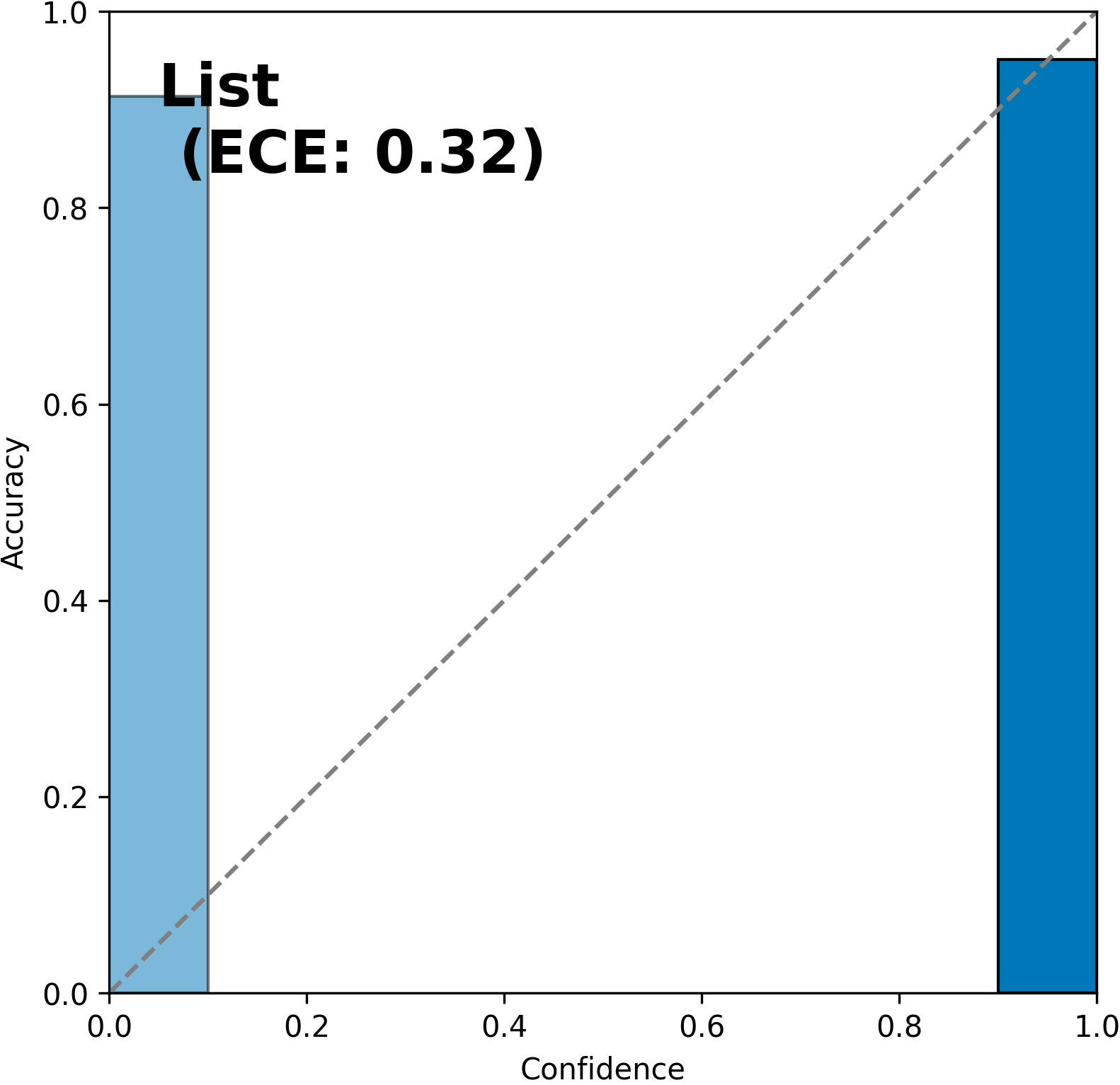}
    \end{subfigure}

    \caption{Reliability diagrams on the development set for eligible methods using Llama3-70B, aggregated across all languages. We use 10 bins in our evaluation. Bar height is the average accuracy of the bin. Darker shades imply higher density of predictions in the bin. }
    \label{fig:calibration_results_llama3_dev}
\end{figure}

\begin{table*}[t]
\footnotesize
\centering
\begin{tabular}{l|cc|cc|cc|cc|cc|cc}
\toprule[0.5pt]
\multicolumn{1}{c}{\multirow{2}{*}{Method}} & \multicolumn{2}{c}{\textbf{En $\rightarrow$ Cs}} & \multicolumn{2}{c}{\textbf{En $\rightarrow$ Hi}} & \multicolumn{2}{c}{\textbf{En $\rightarrow$ Ja}} & \multicolumn{2}{c}{\textbf{En $\rightarrow$ Ru}} & \multicolumn{2}{c}{\textbf{En $\rightarrow$ Zh}} & \multicolumn{2}{c}{\textbf{Average}}\\
\cmidrule(lr){2-3}
\cmidrule(lr){4-5}
\cmidrule(lr){6-7}
\cmidrule(lr){8-9}
\cmidrule(lr){10-11}
\cmidrule(lr){12-13}

& Aya  & Llama  & Aya  & Llama  & Aya  & Llama  & Aya  & Llama  & Aya  & Llama & Aya  & Llama \\
\midrule
Random &        0.07&0.09&0.03&0.03&0.01&0.01&0.08&0.11&0.03&0.04&0.04&0.06     \\
\midrule
Probability&0.17&0.19&0.09&0.08&0.02&0.04&0.19&0.19&\bl{0.11}&0.1&0.12&0.12 \\
Entropy&\bl{0.19}&0.19&0.11&0.09&\bl{0.09}&0.04&\underline{0.19}&0.19&0.08&\underline{0.12}&0.13&0.13 \\
\midrule
List&0.15&0.20&\bl{0.13}&0.08&0.05&0.08&0.18&\bl{0.22}&0.09&0.11&0.12&0.14 \\
Word\_Numeric&\underline{0.18}&0.19&0.12&0.13&\underline{0.06}&\bf{0.13}&\bl{0.21}&0.21&0.09&0.12&0.13&0.16 \\
Word\_Likert&0.12&\bl{0.22}&0.11&\bl{0.15}&0.05&0.10&0.15&\bl{0.23}&0.07&\bl{0.14}&0.10&0.17 \\
Token\_Numeric&0.16&0.19&0.09&0.10&0.06&0.10&0.19&0.21&0.09&0.11&0.12&0.14 \\
Token\_Likert&0.14&0.14&0.09&0.08&0.05&0.09&0.19&0.16&\underline{0.10}&0.08&0.12&0.11 \\
\bottomrule[0.5pt]
\end{tabular}
\caption{F1 scores of negative labels for the binary error detection task on the test set (higher is better). Best performing results are bolded in each column. Underlined values indicate no statistically significant difference from the top score; all other results are significantly lower.}
\label{tab:f1_threshold_all_test_results}
\end{table*}

\begin{table*}
\footnotesize
\centering
\begin{tabular}{l|cc|cc|cc|cc|cc|cc}
\toprule[0.5pt]
\multicolumn{1}{c}{\multirow{2}{*}{Method}} & \multicolumn{2}{c}{\textbf{En $\rightarrow$ Cs}} & \multicolumn{2}{c}{\textbf{En $\rightarrow$ Hi}} & \multicolumn{2}{c}{\textbf{En $\rightarrow$ Ja}} & \multicolumn{2}{c}{\textbf{En $\rightarrow$ Ru}} & \multicolumn{2}{c}{\textbf{En $\rightarrow$ Zh}} & \multicolumn{2}{c}{\textbf{Average}}\\
\cmidrule(lr){2-3}
\cmidrule(lr){4-5}
\cmidrule(lr){6-7}
\cmidrule(lr){8-9}
\cmidrule(lr){10-11}
\cmidrule(lr){12-13}

& Aya  & Llama  & Aya  & Llama  & Aya  & Llama  & Aya  & Llama  & Aya  & Llama & Aya  & Llama \\
Probability&0.12&0.09&0.08&0.06&0.27&0.14&0.13&0.09&0.16&0.11&0.15&0.1 \\
Entropy (normalized) &\bl{0.04}&\bl{0.05}&\bl{0.05}&\bl{0.03}&\bl{0.13}&\bl{0.06}&\bl{0.06}&\bl{0.06}&\bl{0.04}&\bl{0.04}&0.06&0.05\\
\midrule
List & 0.36 & 0.31 & 0.30 & 0.36 & 0.30 & 0.36 & 0.29 & 0.29 & 0.29 & 0.33 & 0.31 & 0.33 \\
Word\_Numeric&0.21&0.22&0.30&0.25&0.34&0.25&0.20&0.21&0.25&0.26&0.26&0.24 \\
Word\_Likert&0.09&0.10&0.13&0.08&\bl{0.13}&0.09&0.13&0.11&0.12&0.08&0.12&0.09 \\
Token\_Numeric&0.47&0.22&0.70&0.21&0.47&0.25&0.40&0.19&0.37&0.30&0.48&0.23 \\
Token\_Likert&0.23&0.10&0.27&0.05&0.23&0.08&0.23&0.10&0.31&0.12&0.25&0.09 \\
\bottomrule[0.5pt]
\end{tabular}
\caption{Results for error calibration on the test set. All the numbers report the ECE score (lower is better). Best performing results are bolded in each column. }
\label{tab:ece_all_test_results}
\end{table*}

\paragraph{Statistical Significance} We assess statistical significance using the bootstrap method with Holm–Bonferroni correction to control the family-wise error rate. Significance is marked in the tables in Section \ref{results}, and full results, including pairwise comparisons between methods, are reported in Appendix \ref{app:significance_testing}.

\section{Results}
\label{results}



\paragraph{F1} The results for error detection are presented in Table \ref{tab:f1_threshold_all_test_results}. The scores are uniformly low but above random, ranging from 0.02 to 0.23. Averaged over each method--model pair, they fall between 0.10 and 0.17. 

For Aya23, Entropy and Word\_Numeric perform best on average. In 3 of the 5 language pairs, they both perform best or are statistically indistinguishable from the best result. For Llama3-70B, Word\_Likert has the best performance in 4 out of 5 language pairs. This shows that no single method consistently outperforms the others; the best-performing method depends on both the language and the model. However, verbalized methods generally perform better with Llama3-70B. The clearest inconsistency appears with Word\_Likert: while it performs best on average for Llama3-70B, it is the worst-performing non-random method for Aya23.

The greatest discrepancy between internal and verbalized methods is seen for English $\rightarrow$ Japanese translations on the Llama3-70B model, where verbalized methods beat internal methods by 0.09 F1. In no setting do Probability and Entropy significantly outperform all verbalized methods. However, verbalized methods outperform them in 5 of the 10 settings. 



\paragraph{ECE} Table \ref{tab:ece_all_test_results} shows the test set results for calibration. Scores span from 0.04–0.70, with method–model pair averages ranging from 0.05 to 0.48.  For both models, the normalized Entropy consistently yields the best results. For Aya23, Word\_Likert is the second best method on average, being on par with Entropy for English-to-Japanese translations. 
For Llama3-70B, Probability, Word\_Likert and Token\_Likert show comparable results on average.  

We use reliability diagrams to visualize calibration. They partition predictions into bins according to their confidence, and showcase the average accuracy of each bin \cite{niculescu2005predicting}.  
Figure \ref{fig:calibration_results_llama3_dev} shows the reliability diagrams for the aggregation of all tokens in the development set for the Llama3-70B model. As we can see, all methods \textit{underestimate} the correctness of miscalibrated tokens.

Likert methods consistently outperform their Numeric counterparts in terms of calibration. The weak calibration of numeric methods is consistent with prior work; LLMs generally show poor calibration in numerical contexts \cite{lovering-etal-2025-language}.

Interestingly, even though Token\_Likert is one of the best performing methods for Llama3-70B, it does \textit{not} perform competitively for Aya23. Word\_Likert is more consistent on both models. This can perhaps be explained in part by differences in tokenization of the two models, while the word segmentation process used for Word\_Numeric and Word\_Likert is model-agnostic.

\paragraph{AUROC and AUPRC} Tables \ref{tab:auroc_all_test_results} and \ref{tab:average_precision_all_test_results} (in the appendix) report AUROC and AUPRC results, respectively. 
We observe mixed results overall, with generally weak alignment across methods. For AUROC, Entropy consistently performs best for Aya23, while for Llama3-70B, Word\_Numeric and Token\_Numeric outperform other measures. For AUPRC, internal measures generally yield better results than verbalized ones for Aya23, whereas for Llama3-70B, Word\_Numeric and Word\_Likert perform on par with internal methods on average. These results are largely consistent with those for the F1 score.





\begin{table*}
\footnotesize
\centering
{
\setlength{\tabcolsep}{4.5pt}
\begin{tabular}{cl|cc|cc|cc|cc|cc|cc}
\toprule[0.5pt]
& \multicolumn{1}{c}{\multirow{2}{*}{Method}} & \multicolumn{2}{c}{\textbf{En $\rightarrow$ Cs}} & \multicolumn{2}{c}{\textbf{En $\rightarrow$ Hi}} & \multicolumn{2}{c}{\textbf{En $\rightarrow$ Ja}} & \multicolumn{2}{c}{\textbf{En $\rightarrow$ Ru}} & \multicolumn{2}{c}{\textbf{En $\rightarrow$ Zh}} & \multicolumn{2}{c}{\textbf{Average}}\\
\cmidrule(lr){3-4}
\cmidrule(lr){5-6}
\cmidrule(lr){7-8}
\cmidrule(lr){9-10}
\cmidrule(lr){11-12}
\cmidrule(lr){13-14}

& & Aya  & Llama  & Aya  & Llama  & Aya  & Llama  & Aya  & Llama  & Aya  & Llama & Aya  & Llama \\

\multirow{5}{*}{\rotatebox{90}{Probability}}
&List&0.05&\bl{0.15}&0.02&0.07&0.06&0.10&0.05&0.14&0.03&0.09&0.04&0.11 \\
&Word\_Numeric&\bl{0.14}&\bl{0.15}&\bl{0.10}&\bl{0.13}&\bl{0.09}&\bl{0.17}&\bl{0.14}&\bl{0.16}&\bl{0.17}&\bl{0.18}&0.13&0.16 \\
&Word\_Likert&0.06&0.13&0.06&0.11&0.03&0.12&0.04&0.14&0.05&0.12&0.05&0.12 \\
&Token\_Numeric&-0.14&0.09&-0.12&0.05&-0.06&0.07&-0.12&0.13&0.08&0.05&-0.07&0.08 \\
&Token\_Likert&-0.09&0.00&0.01&-0.03&-0.02&-0.09&-0.08&0.05&0.03&-0.11&-0.03&-0.04 \\
\midrule
\multirow{5}{*}{\rotatebox{90}{Entropy}}
&List&0.05&\bl{0.15}&0.02&0.07&0.06&0.11&0.06&0.14&0.04&0.10&0.05&0.11\\
&Word\_Numeric&\bl{0.14}&\bl{0.16}&\bl{0.10}&\bl{0.14}&0.09&\bl{0.18}&\bl{0.14}&\bl{0.17}&\bl{0.18}&\bl{0.19}&0.13&0.17 \\
&Word\_Likert&0.06&0.13&0.06&0.12&\bl{0.11}&0.13&0.04&0.14&0.06&0.13&0.05&0.13 \\
&Token\_Numeric&-0.13&0.09&-0.12&0.05&-0.08&0.07&-0.12&0.13&0.09&0.06&-0.07&0.08 \\
&Token\_Likert&-0.09&0.00&0.00&-0.03&-0.05&-0.08&-0.09&0.05&0.03&-0.11&-0.04&-0.03 \\

\bottomrule[0.5pt]
\end{tabular}
}
\caption{Results for the alignment of continuous verbalized methods with token probabilities (top four rows) and with token entropies (bottom four rows). All the numbers report Spearman's correlation coefficient (higher indicates higher alignment). Best alignment with either probabilities or entropies are bolded in each column.}
\label{tab:spearman_continuous_test_results}
\end{table*}






\section{Analysis}
\label{discussion}


\subsection{Alignment of Internal and Verbalized Measures} 
Given the similar performance of internal and verbalized measures, we further ask if there is substantial agreement between the two types of measures. 
Therefore, we evaluate the correlation of verbalized and internal measures with each other. Following \citet{kumar-etal-2024-confidence}, we report \bl{Spearman's rank correlation coefficient}. Given variables $X$ and $Y,$ each of size $n$, we have $n$ pairs of raw scores $(X_i, Y_i)$. Spearman's coefficient is calculated as follows:
\[
\rho = 1 - \frac{6 \sum d_i^2}{n(n^2 - 1)}
\]
where $d_i$ is the difference in \textit{rank} of $X_i$ and $Y_i$ in their corresponding lists.

Tables \ref{tab:spearman_continuous_test_results} (next page) and \ref{tab:list_verbalized_internal_f1_test_results} and \ref{tab:list_verbalized_internal_auroc_test_results} (in the appendix) show the alignment of verbalized confidence and internal certainty. Word\_Numeric yields the highest correlation with probability in all 10 settings and the highest correlation with entropy in 9 of the 10 settings. Overall, however, we see that, for the two models in our experiments, there is little to no correlation between internal and verbalized measures. Comparing the two models, Llama3-70B generally achieves slightly higher correlations than Aya23. Appendix \ref{app:verbal_internal} presents an example of the disagreement between internal and verbalized measures. 

\begin{table}[t]
\footnotesize
\centering
\begin{tabular}{l|c|c|c|c}
\toprule[0.5pt]
\multicolumn{1}{c}{\bl{Method}} & \multicolumn{1}{c}{\textbf{1}} & 
\multicolumn{1}{c}{\textbf{0.5}}&
\multicolumn{1}{c}{\textbf{0.2}} &
\multicolumn{1}{c}{\textbf{0.1}} \\
\midrule
Word\_Numeric & 0.159 & 0.162 & 0.968 & 1 \\
Token\_Numeric & 0.542 & 0.542 & 0.99 & 1\\
\midrule
Word\_Numeric & 0.028& 0.028& 0.493& 1 \\
Token\_Numeric & 0.027 & 0.027 & 0.471&  1\\
\bottomrule[0.5pt]
\end{tabular}
\caption{The ratio of numerical confidence outputs that are multiples of the respective column number, as aggregated across all languages. All columns include the number 0 as a multiple. The results are for the test set, for Aya23 (top two rows) and Llama3-70B (bottom two rows).}
\label{tab:numerical_ratios}
\end{table}

\subsection{Precision and Recall Tradeoff}

\begin{figure}[H]
    \centering
     \includegraphics[width=1\linewidth]{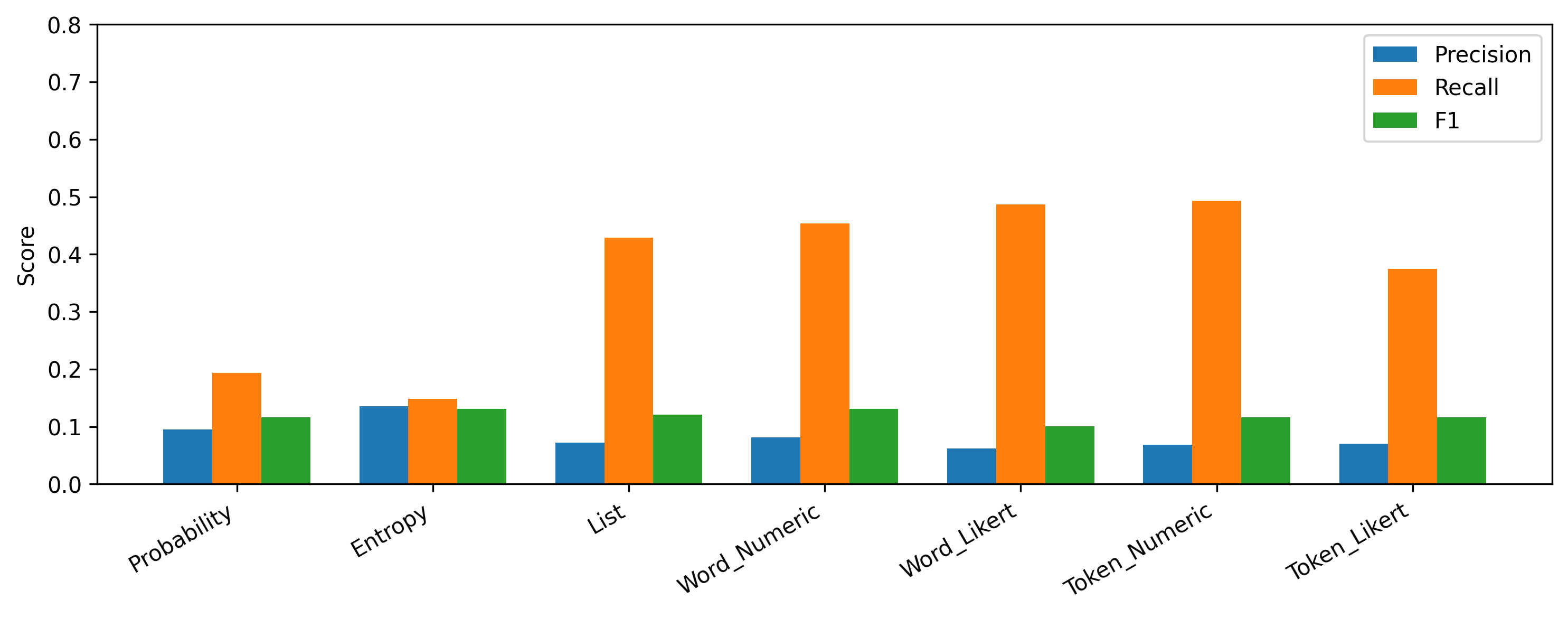} \\
    \includegraphics[width=1\linewidth]{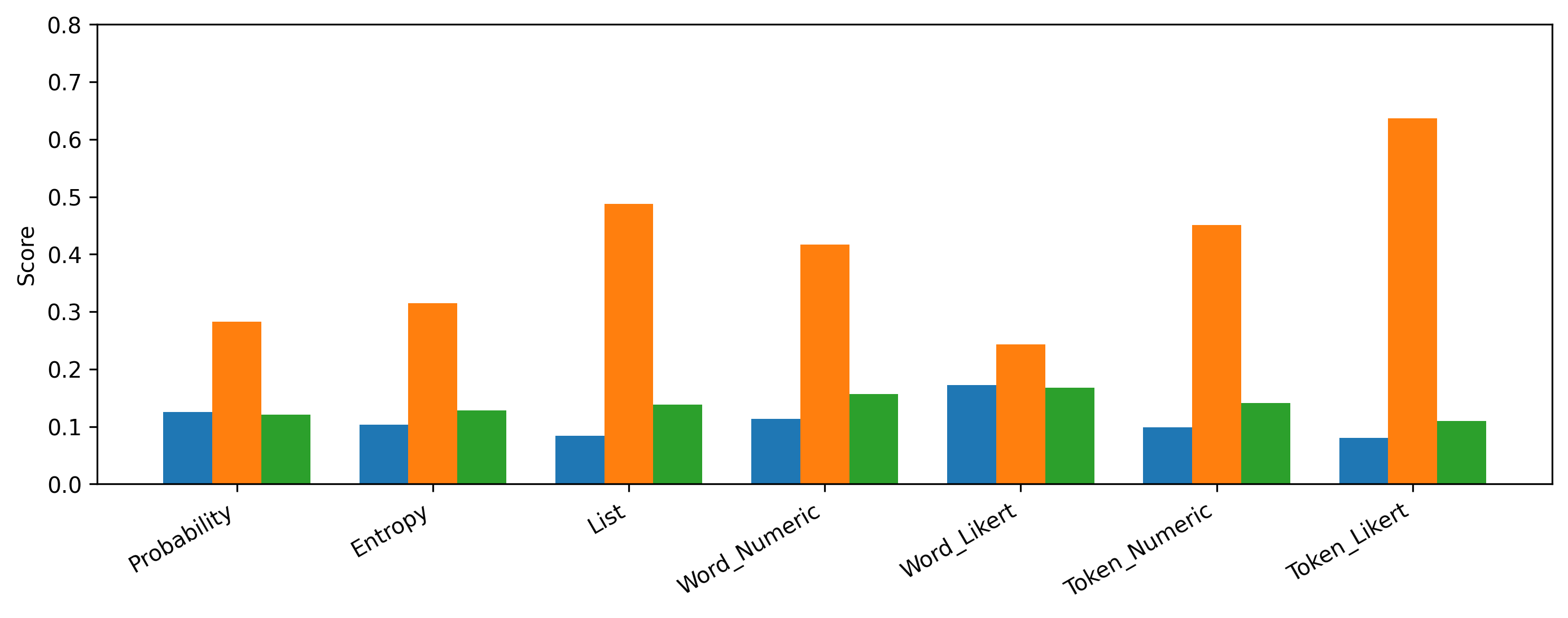}
    \caption{F1, precision, and recall score for different methods averaged over all the languages on the test set for Aya23 (top) and Llama3-70B (bottom). }
    \label{fig:f1_pre_rec_llama_test}
\end{figure}

Figure \ref{fig:f1_pre_rec_llama_test} shows the F1, precision, and recall for different methods averaged across all languages for Aya23 and Llama3-70B. Verbalized methods typically have higher recalls than internal methods. Verbalized methods have lower precision for Aya23 but they match internal methods for Llama3-70B, particularly for the Word\_Likert and Word\_Numeric methods. 

Some methods behave noticeably differently between the two models. For Llama3-70B, Word\_Numeric has higher recall than Word\_Likert, and Token\_Numeric has lower recall than Token\_Likert. For Aya23, both relations are reversed. Notably, the List method consistently has some of the highest recalls of all methods in both models. In contrast, Word\_Likert has a very high recall for Aya23 but the lowest recall for Llama3-70B.

\subsection{Granularity}
 One advantage that verbalized Numeric methods might have over their Likert counterparts is their ability to return arbitrarily fine-grained values. However, from Figure \ref{fig:side_by_side_pre_rec_cur_russian} (in the appendix) it seems that, in practice, the two models do not make use of this flexibility. Table \ref{tab:numerical_ratios} shows the ratios of different levels of granularity for the numeric methods for both models. For Aya23, the confidences are rarely \textit{not} multiples of $0.2$. For both models, all confidences are multiples of $0.1$. Thus, even the Numeric methods are limited to at most 11 confidence levels (multiples of 0.1 in the range $[0, 1]$), compared to the 6 levels of Likert methods.


\begin{figure}[t]
    \centering
    \includegraphics[width=1\linewidth]{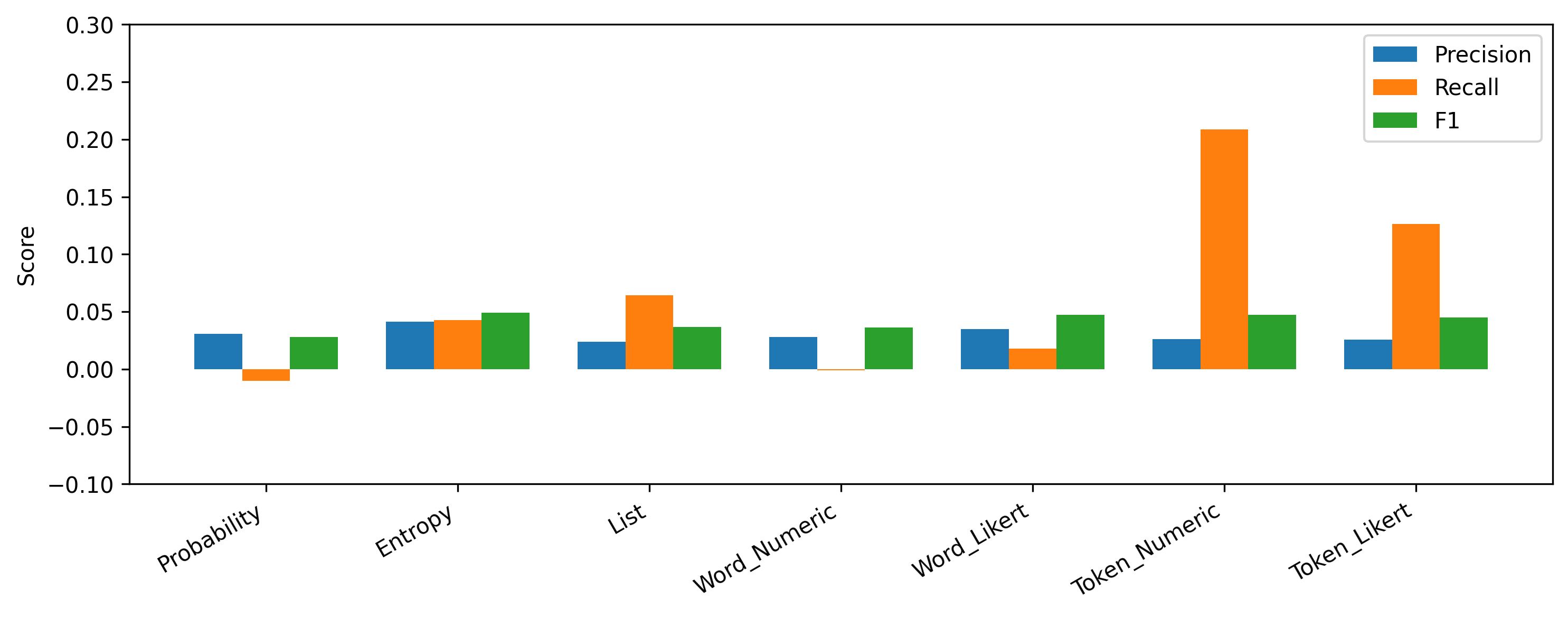} \\
    \includegraphics[width=1\linewidth]{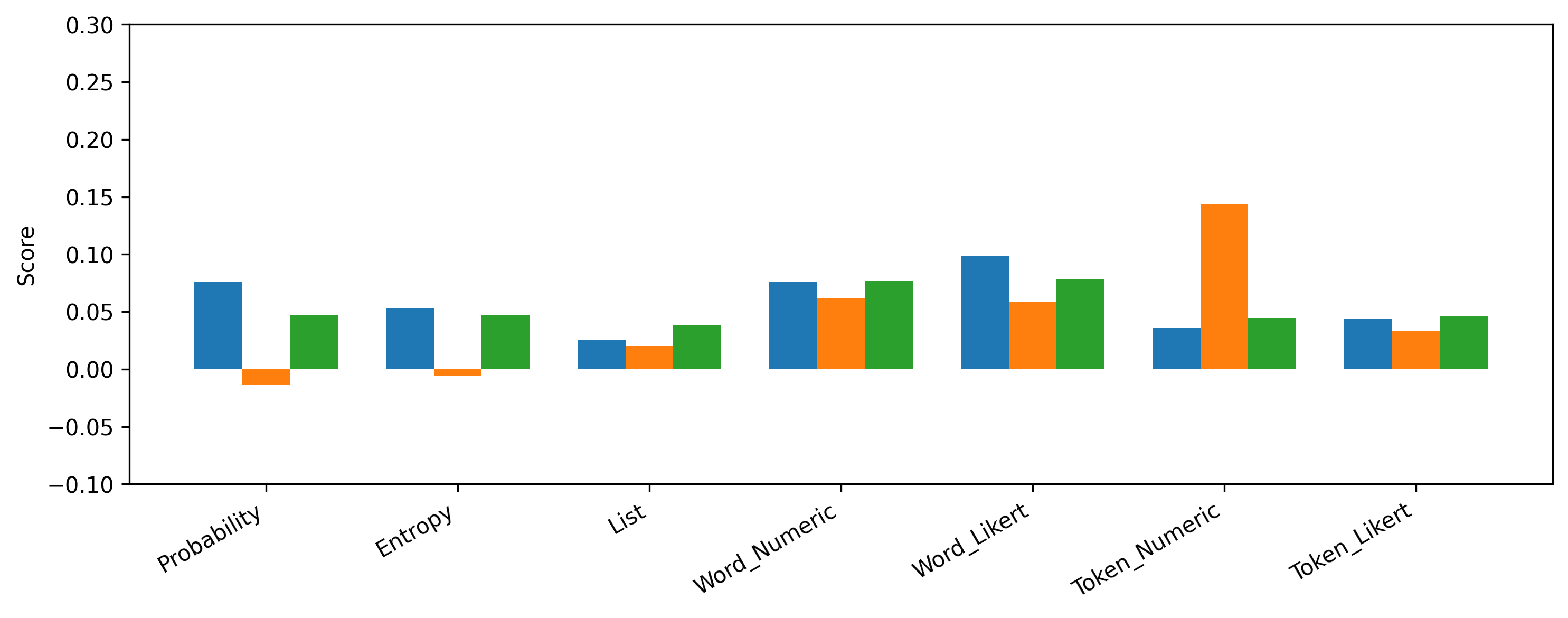}
    \caption{The difference between average precision, recall, and F1 for open vs closed class words on the test set for Aya23 (top) and Llama3-70B (bottom). Each column denotes the performance for open tokens minus that of closed tokens.}
    \label{fig:open_vs_closed}
\end{figure}


\subsection{Syntactic Categories} To see differences between the methods, we use grammatical functions as our main point of comparison. We use Stanza \cite{qi-etal-2020-stanza} for part-of-speech tagging of the dataset. For the purpose of our analysis, we assign the syntactic category of a word to each of its constituent tokens.

Figures \ref{fig:aya_pos_piechart1} and \ref{fig:llama_pos_piechart} (in the appendix) show, for each method and model, the distribution of the POS tags in the test set that the model labels as \textit{incorrect}. Additionally, we report the Hellinger distance\footnote{For two discrete probability distributions, Hellinger distance is defined as the Euclidean distance between their element-wise square roots.} of these distributions from that of the real errors. For both models Token\_Numeric, Token\_Likert and List are consistently the closest in distribution to the real error distribution. 
For both models, Word\_Numeric and Word\_Likert report an unusually high proportion of punctuation errors, and a lower-than-expected proportion for nouns and proper nouns.

Alternatively, each word belongs to either an open class or a closed class.
Open classes typically allow new words to be added, while closed classes rarely gain new members. The open classes typically contain words with richer semantic content (like nouns, verbs, and adjectives). The closed classes are usually functional words serving grammatical purposes (like adpositions or articles).\footnote{Note that this is not universal. The distinction of open and closed classes is often compared to that of content and function words, but that is also debated.}

Figure \ref{fig:open_vs_closed} shows the average difference of performance for open and closed class tokens on the test set. For both models and all methods, we see higher average F1 and precision for open class words.
From Section \ref{results}, we know that verbalized methods generally perform better with Llama3-70B. Here, they show higher recall for open class words than for closed class words, whereas internal methods do not differentiate in recall. For Aya23, verbalized methods do not outperform internal methods, as shown in Section \ref{results}. They also do not show the same recall pattern observed for Llama3-70B. Appendix \ref{sec:app_pos_specific_charts} presents the results for the performance of all methods on each ``open'' syntactic category, along with some additional analysis.

\section{Conclusion and Future Work}

For MT error detection, traditional unsupervised techniques that utilize the model's internal representations, such as predicted probabilities, can be flawed because they reflect certainty among competing  hypotheses rather than correctness.
We investigate verbalized confidence as an alternative unsupervised approach.
For both binary error detection and calibration, we show that verbalized confidence performs comparably to internal methods.  
The List method and the word-based methods also have the advantage of being more easily accessible without requiring model transparency.
We note that the performance of all methods, whether verbalized or internal, remains relatively low. 
This is partly due to the difficulty of the task, as translation quality is typically very high for high-resource languages.
As for the alignment between internal uncertainty and verbalized confidence, we find  little to no correlation between them for either model, although Llama3-70B shows slightly higher correlations than Aya23. Notably, Llama3-70B also derives greater benefits from verbalized measures in terms of alignment with ground truth.
Future work is needed to identify the factors that affect these alignments.
 
Poorer MT performance in low-resource languages creates an even greater need for quality control. Future work should examine the extent to which our findings generalize to lower-resource scenarios. Although using Translation Edit Rate to generate pseudo-labels may seem like an expedient solution in the absence of human annotations, such labels are not reliably consistent with human-annotated labels. Therefore, we emphasize the necessity of human annotation for these languages over the use of automated proxies.

\section*{Limitations}

The main limitation of our work is the scarce availability of annotated translations for LLM outputs. Such annotations are rare even for traditional, non-LLM MT systems. Therefore, we were unable to 
extend our study to a wider range of morphologically diverse languages, despite the potential benefits for low-resource settings.

Although some of the available annotations are available for closed-source LLMs in the WMT24 dataset, we exclude these models from experiments  due to cost constraints.

\section*{Ethical Considerations}

These unsupervised automatic error detection methods generally show limited effectiveness and are not a replacement for human annotation.

We adhere to the licensing agreement of Llama3-70B\footnote{\url{https://www.llama.com/llama3/license/}} and Aya23\footnote{CC-BY-NC 4.0\url{}}. 
The dataset we use in our work from \citet{kocmi-etal-2024-findings} is already fully pseudonymized (using random first names and surnames) and the required steps were taken to ensure safety and anonymity.

Additionally, our work relied on significant computational resources, with unavoidable environmental impact.


\bibliography{custom}

\begin{thebibliography}{44}
\providecommand{\natexlab}[1]{#1}

\bibitem[{Aryabumi et~al.(2024)Aryabumi, Dang, Talupuru, Dash, Cairuz, Lin, Venkitesh, Smith, Campos, Tan et~al.}]{aryabumi2024aya}
Viraat Aryabumi, John Dang, Dwarak Talupuru, Saurabh Dash, David Cairuz, Hangyu Lin, Bharat Venkitesh, Madeline Smith, Jon~Ander Campos, Yi~Chern Tan, and 1 others. 2024.
\newblock Aya 23: Open weight releases to further multilingual progress.
\newblock \emph{arXiv preprint arXiv:2405.15032}.

\bibitem[{Berg-Kirkpatrick et~al.(2012)Berg-Kirkpatrick, Burkett, and Klein}]{berg-kirkpatrick-etal-2012-empirical}
Taylor Berg-Kirkpatrick, David Burkett, and Dan Klein. 2012.
\newblock \href {https://aclanthology.org/D12-1091/} {An empirical investigation of statistical significance in {NLP}}.
\newblock In \emph{Proceedings of the 2012 Joint Conference on Empirical Methods in Natural Language Processing and Computational Natural Language Learning}, pages 995--1005, Jeju Island, Korea. Association for Computational Linguistics.

\bibitem[{Dinh and Niehues(2025)}]{dinh-niehues-2025-generative}
Tu~Anh Dinh and Jan Niehues. 2025.
\newblock \href {https://doi.org/10.18653/v1/2025.emnlp-main.166} {Are generative models underconfident? better quality estimation with boosted model probability}.
\newblock In \emph{Proceedings of the 2025 Conference on Empirical Methods in Natural Language Processing}, pages 3364--3382, Suzhou, China. Association for Computational Linguistics.

\bibitem[{Fomicheva et~al.(2022)Fomicheva, Sun, Fonseca, Zerva, Blain, Chaudhary, Guzm{\'a}n, Lopatina, Specia, and Martins}]{fomicheva-etal-2022-mlqe}
Marina Fomicheva, Shuo Sun, Erick Fonseca, Chrysoula Zerva, Fr{\'e}d{\'e}ric Blain, Vishrav Chaudhary, Francisco Guzm{\'a}n, Nina Lopatina, Lucia Specia, and Andr{\'e} F.~T. Martins. 2022.
\newblock \href {https://aclanthology.org/2022.lrec-1.530/} {{MLQE}-{PE}: A multilingual quality estimation and post-editing dataset}.
\newblock In \emph{Proceedings of the Thirteenth Language Resources and Evaluation Conference}, pages 4963--4974, Marseille, France. European Language Resources Association.

\bibitem[{Gou et~al.(2023)Gou, Shao, Gong, Shen, Yang, Duan, and Chen}]{gou2023critic}
Zhibin Gou, Zhihong Shao, Yeyun Gong, Yelong Shen, Yujiu Yang, Nan Duan, and Weizhu Chen. 2023.
\newblock Critic: Large language models can self-correct with tool-interactive critiquing.
\newblock \emph{arXiv preprint arXiv:2305.11738}.

\bibitem[{Grattafiori et~al.(2024)Grattafiori, Dubey, Jauhri, Pandey, Kadian, Al-Dahle, Letman, Mathur, Schelten, Vaughan et~al.}]{grattafiori2024llama}
Aaron Grattafiori, Abhimanyu Dubey, Abhinav Jauhri, Abhinav Pandey, Abhishek Kadian, Ahmad Al-Dahle, Aiesha Letman, Akhil Mathur, Alan Schelten, Alex Vaughan, and 1 others. 2024.
\newblock The llama 3 herd of models.
\newblock \emph{arXiv preprint arXiv:2407.21783}.

\bibitem[{Han et~al.(2022)Han, Hao, Dong, Sun, and Wei}]{han2022prototypical}
Zhixiong Han, Yaru Hao, Li~Dong, Yutao Sun, and Furu Wei. 2022.
\newblock Prototypical calibration for few-shot learning of language models.
\newblock \emph{arXiv preprint arXiv:2205.10183}.

\bibitem[{Holtzman et~al.(2021)Holtzman, West, Shwartz, Choi, and Zettlemoyer}]{holtzman-etal-2021-surface}
Ari Holtzman, Peter West, Vered Shwartz, Yejin Choi, and Luke Zettlemoyer. 2021.
\newblock \href {https://doi.org/10.18653/v1/2021.emnlp-main.564} {Surface form competition: Why the highest probability answer isn{'}t always right}.
\newblock In \emph{Proceedings of the 2021 Conference on Empirical Methods in Natural Language Processing}, pages 7038--7051, Online and Punta Cana, Dominican Republic. Association for Computational Linguistics.

\bibitem[{Kadavath et~al.(2022)Kadavath, Conerly, Askell, Henighan, Drain, Perez, Schiefer, Hatfield-Dodds, DasSarma, Tran-Johnson et~al.}]{kadavath2022language}
Saurav Kadavath, Tom Conerly, Amanda Askell, Tom Henighan, Dawn Drain, Ethan Perez, Nicholas Schiefer, Zac Hatfield-Dodds, Nova DasSarma, Eli Tran-Johnson, and 1 others. 2022.
\newblock Language models (mostly) know what they know.
\newblock \emph{arXiv preprint arXiv:2207.05221}.

\bibitem[{Kocmi et~al.(2025)Kocmi, Artemova, Avramidis, Bawden, Bojar, Dranch, Dvorkovich, Dukanov, Fishel, Freitag, Gowda, Grundkiewicz, Haddow, Karpinska, Koehn, Lakougna, Lundin, Monz, Murray, Nagata, Perrella, Proietti, Popel, Popovi{\'c}, Riley, Shmatova, Steingr{\'i}msson, Yankovskaya, and Zouhar}]{kocmi-etal-2025-findings}
Tom Kocmi, Ekaterina Artemova, Eleftherios Avramidis, Rachel Bawden, Ond{\v{r}}ej Bojar, Konstantin Dranch, Anton Dvorkovich, Sergey Dukanov, Mark Fishel, Markus Freitag, Thamme Gowda, Roman Grundkiewicz, Barry Haddow, Marzena Karpinska, Philipp Koehn, Howard Lakougna, Jessica Lundin, Christof Monz, Kenton Murray, and 10 others. 2025.
\newblock \href {https://doi.org/10.18653/v1/2025.wmt-1.22} {Findings of the {WMT}25 general machine translation shared task: Time to stop evaluating on easy test sets}.
\newblock In \emph{Proceedings of the Tenth Conference on Machine Translation}, pages 355--413, Suzhou, China. Association for Computational Linguistics.

\bibitem[{Kocmi et~al.(2024{\natexlab{a}})Kocmi, Avramidis, Bawden, Bojar, Dvorkovich, Federmann, Fishel, Freitag, Gowda, Grundkiewicz, Haddow, Karpinska, Koehn, Marie, Monz, Murray, Nagata, Popel, Popovi{\'c}, Shmatova, Steingr{\'i}msson, and Zouhar}]{kocmi-etal-2024-findings}
Tom Kocmi, Eleftherios Avramidis, Rachel Bawden, Ond{\v{r}}ej Bojar, Anton Dvorkovich, Christian Federmann, Mark Fishel, Markus Freitag, Thamme Gowda, Roman Grundkiewicz, Barry Haddow, Marzena Karpinska, Philipp Koehn, Benjamin Marie, Christof Monz, Kenton Murray, Masaaki Nagata, Martin Popel, Maja Popovi{\'c}, and 3 others. 2024{\natexlab{a}}.
\newblock \href {https://doi.org/10.18653/v1/2024.wmt-1.1} {Findings of the {WMT}24 general machine translation shared task: The {LLM} era is here but {MT} is not solved yet}.
\newblock In \emph{Proceedings of the Ninth Conference on Machine Translation}, pages 1--46, Miami, Florida, USA. Association for Computational Linguistics.

\bibitem[{Kocmi et~al.(2024{\natexlab{b}})Kocmi, Zouhar, Avramidis, Grundkiewicz, Karpinska, Popovi{\'c}, Sachan, and Shmatova}]{kocmi-etal-2024-error}
Tom Kocmi, Vil{\'e}m Zouhar, Eleftherios Avramidis, Roman Grundkiewicz, Marzena Karpinska, Maja Popovi{\'c}, Mrinmaya Sachan, and Mariya Shmatova. 2024{\natexlab{b}}.
\newblock \href {https://doi.org/10.18653/v1/2024.wmt-1.131} {Error span annotation: A balanced approach for human evaluation of machine translation}.
\newblock In \emph{Proceedings of the Ninth Conference on Machine Translation}, pages 1440--1453, Miami, Florida, USA. Association for Computational Linguistics.

\bibitem[{Kumar et~al.(2024)Kumar, Morabito, Umbet, Kabbara, and Emami}]{kumar-etal-2024-confidence}
Abhishek Kumar, Robert Morabito, Sanzhar Umbet, Jad Kabbara, and Ali Emami. 2024.
\newblock \href {https://doi.org/10.18653/v1/2024.acl-long.20} {Confidence under the hood: An investigation into the confidence-probability alignment in large language models}.
\newblock In \emph{Proceedings of the 62nd Annual Meeting of the Association for Computational Linguistics (Volume 1: Long Papers)}, pages 315--334, Bangkok, Thailand. Association for Computational Linguistics.

\bibitem[{Kumar and Sarawagi(2019)}]{kumar2019calibration}
Aviral Kumar and Sunita Sarawagi. 2019.
\newblock Calibration of encoder decoder models for neural machine translation.
\newblock \emph{arXiv preprint arXiv:1903.00802}.

\bibitem[{Lin et~al.(2022)Lin, Hilton, and Evans}]{lin2022teaching}
Stephanie Lin, Jacob Hilton, and Owain Evans. 2022.
\newblock Teaching models to express their uncertainty in words.
\newblock \emph{arXiv preprint arXiv:2205.14334}.

\bibitem[{Liu et~al.(2025)Liu, Chen, Da, Chen, Lin, and Wei}]{liu2025uncertainty}
Xiaoou Liu, Tiejin Chen, Longchao Da, Chacha Chen, Zhen Lin, and Hua Wei. 2025.
\newblock Uncertainty quantification and confidence calibration in large language models: A survey.
\newblock In \emph{Proceedings of the 31st ACM SIGKDD Conference on Knowledge Discovery and Data Mining V. 2}, pages 6107--6117.

\bibitem[{Lovering et~al.(2024)Lovering, Krumdick, Lai, Ebner, Kumar, Reddy, Koncel-Kedziorski, and Tanner}]{lovering2024language}
Charles Lovering, Michael Krumdick, Viet~Dac Lai, Seth Ebner, Nilesh Kumar, Varshini Reddy, Rik Koncel-Kedziorski, and Chris Tanner. 2024.
\newblock Language model probabilities are not calibrated in numeric contexts.
\newblock \emph{arXiv preprint arXiv:2410.16007}.

\bibitem[{Lovering et~al.(2025)Lovering, Krumdick, Lai, Reddy, Ebner, Kumar, Koncel-Kedziorski, and Tanner}]{lovering-etal-2025-language}
Charles Lovering, Michael Krumdick, Viet~Dac Lai, Varshini Reddy, Seth Ebner, Nilesh Kumar, Rik Koncel-Kedziorski, and Chris Tanner. 2025.
\newblock \href {https://doi.org/10.18653/v1/2025.acl-long.1417} {Language model probabilities are $not$ calibrated in numeric contexts}.
\newblock In \emph{Proceedings of the 63rd Annual Meeting of the Association for Computational Linguistics (Volume 1: Long Papers)}, pages 29218--29257, Vienna, Austria. Association for Computational Linguistics.

\bibitem[{Lu et~al.(2022)Lu, Zeng, Zhang, Wu, and Li}]{lu-etal-2022-learning}
Yu~Lu, Jiali Zeng, Jiajun Zhang, Shuangzhi Wu, and Mu~Li. 2022.
\newblock \href {https://doi.org/10.18653/v1/2022.acl-long.167} {Learning confidence for transformer-based neural machine translation}.
\newblock In \emph{Proceedings of the 60th Annual Meeting of the Association for Computational Linguistics (Volume 1: Long Papers)}, pages 2353--2364, Dublin, Ireland. Association for Computational Linguistics.

\bibitem[{Manakul et~al.(2023)Manakul, Liusie, and Gales}]{manakul-etal-2023-selfcheckgpt}
Potsawee Manakul, Adian Liusie, and Mark Gales. 2023.
\newblock \href {https://doi.org/10.18653/v1/2023.emnlp-main.557} {{S}elf{C}heck{GPT}: Zero-resource black-box hallucination detection for generative large language models}.
\newblock In \emph{Proceedings of the 2023 Conference on Empirical Methods in Natural Language Processing}, pages 9004--9017, Singapore. Association for Computational Linguistics.

\bibitem[{Mielke et~al.(2022)Mielke, Szlam, Dinan, and Boureau}]{mielke-etal-2022-reducing}
Sabrina~J. Mielke, Arthur Szlam, Emily Dinan, and Y-Lan Boureau. 2022.
\newblock \href {https://doi.org/10.1162/tacl_a_00494} {Reducing conversational agents' overconfidence through linguistic calibration}.
\newblock \emph{Transactions of the Association for Computational Linguistics}, 10:857--872.

\bibitem[{Naeini et~al.(2015)Naeini, Cooper, and Hauskrecht}]{naeini2015obtaining}
Mahdi~Pakdaman Naeini, Gregory Cooper, and Milos Hauskrecht. 2015.
\newblock Obtaining well calibrated probabilities using bayesian binning.
\newblock In \emph{Proceedings of the AAAI conference on artificial intelligence}, volume~29.

\bibitem[{Ni et~al.(2024)Ni, Bi, Yu, and Guo}]{ni2024large}
Shiyu Ni, Keping Bi, Lulu Yu, and Jiafeng Guo. 2024.
\newblock Are large language models more honest in their probabilistic or verbalized confidence?
\newblock In \emph{China Conference on Information Retrieval}, pages 124--135. Springer.

\bibitem[{Niculescu-Mizil and Caruana(2005)}]{niculescu2005predicting}
Alexandru Niculescu-Mizil and Rich Caruana. 2005.
\newblock Predicting good probabilities with supervised learning.
\newblock In \emph{Proceedings of the 22nd international conference on Machine learning}, pages 625--632.

\bibitem[{Ott et~al.(2018)Ott, Auli, Grangier, and Ranzato}]{ott2018analyzing}
Myle Ott, Michael Auli, David Grangier, and Marc’Aurelio Ranzato. 2018.
\newblock Analyzing uncertainty in neural machine translation.
\newblock In \emph{International Conference on Machine Learning}, pages 3956--3965. PMLR.

\bibitem[{Qi et~al.(2020)Qi, Zhang, Zhang, Bolton, and Manning}]{qi-etal-2020-stanza}
Peng Qi, Yuhao Zhang, Yuhui Zhang, Jason Bolton, and Christopher~D. Manning. 2020.
\newblock \href {https://doi.org/10.18653/v1/2020.acl-demos.14} {{S}tanza: A python natural language processing toolkit for many human languages}.
\newblock In \emph{Proceedings of the 58th Annual Meeting of the Association for Computational Linguistics: System Demonstrations}, pages 101--108, Online. Association for Computational Linguistics.

\bibitem[{Qi et~al.(2021)Qi, Luo, Xu, Ji, and Yang}]{qi2021stochastic}
Qi~Qi, Youzhi Luo, Zhao Xu, Shuiwang Ji, and Tianbao Yang. 2021.
\newblock Stochastic optimization of areas under precision-recall curves with provable convergence.
\newblock \emph{Advances in neural information processing systems}, 34:1752--1765.

\bibitem[{Sarti et~al.(2022)Sarti, Bisazza, Guerberof-Arenas, and Toral}]{sarti-etal-2022-divemt}
Gabriele Sarti, Arianna Bisazza, Ana Guerberof-Arenas, and Antonio Toral. 2022.
\newblock \href {https://doi.org/10.18653/v1/2022.emnlp-main.532} {{D}iv{EMT}: Neural machine translation post-editing effort across typologically diverse languages}.
\newblock In \emph{Proceedings of the 2022 Conference on Empirical Methods in Natural Language Processing}, pages 7795--7816, Abu Dhabi, United Arab Emirates. Association for Computational Linguistics.

\bibitem[{Sarti et~al.(2025{\natexlab{a}})Sarti, Zouhar, Chrupa{\l}a, Guerberof-Arenas, Nissim, and Bisazza}]{sarti2025qe4pe}
Gabriele Sarti, Vil{\'e}m Zouhar, Grzegorz Chrupa{\l}a, Ana Guerberof-Arenas, Malvina Nissim, and Arianna Bisazza. 2025{\natexlab{a}}.
\newblock Qe4pe: Word-level quality estimation for human post-editing.
\newblock \emph{Transactions of the Association for Computational Linguistics}, 13:1410--1435.

\bibitem[{Sarti et~al.(2025{\natexlab{b}})Sarti, Zouhar, Nissim, and Bisazza}]{sarti-etal-2025-unsupervised}
Gabriele Sarti, Vil{\'e}m Zouhar, Malvina Nissim, and Arianna Bisazza. 2025{\natexlab{b}}.
\newblock \href {https://doi.org/10.18653/v1/2025.emnlp-main.924} {Unsupervised word-level quality estimation for machine translation through the lens of annotators (dis)agreement}.
\newblock In \emph{Proceedings of the 2025 Conference on Empirical Methods in Natural Language Processing}, pages 18320--18337, Suzhou, China. Association for Computational Linguistics.

\bibitem[{Snover et~al.(2006)Snover, Dorr, Schwartz, Micciulla, and Makhoul}]{snover-etal-2006-study}
Matthew Snover, Bonnie Dorr, Rich Schwartz, Linnea Micciulla, and John Makhoul. 2006.
\newblock \href {https://aclanthology.org/2006.amta-papers.25/} {A study of translation edit rate with targeted human annotation}.
\newblock In \emph{Proceedings of the 7th Conference of the Association for Machine Translation in the Americas: Technical Papers}, pages 223--231, Cambridge, Massachusetts, USA. Association for Machine Translation in the Americas.

\bibitem[{Tian et~al.(2023)Tian, Mitchell, Zhou, Sharma, Rafailov, Yao, Finn, and Manning}]{tian-etal-2023-just}
Katherine Tian, Eric Mitchell, Allan Zhou, Archit Sharma, Rafael Rafailov, Huaxiu Yao, Chelsea Finn, and Christopher Manning. 2023.
\newblock \href {https://doi.org/10.18653/v1/2023.emnlp-main.330} {Just ask for calibration: Strategies for eliciting calibrated confidence scores from language models fine-tuned with human feedback}.
\newblock In \emph{Proceedings of the 2023 Conference on Empirical Methods in Natural Language Processing}, pages 5433--5442, Singapore. Association for Computational Linguistics.

\bibitem[{Tsai et~al.(2024)Tsai, Talbott, and Zhang}]{tsai2024efficient}
Yao-Hung~Hubert Tsai, Walter Talbott, and Jian Zhang. 2024.
\newblock Efficient non-parametric uncertainty quantification for black-box large language models and decision planning.
\newblock \emph{arXiv preprint arXiv:2402.00251}.

\bibitem[{Ulmer et~al.(2024)Ulmer, Gubri, Lee, Yun, and Oh}]{ulmer-etal-2024-calibrating}
Dennis Ulmer, Martin Gubri, Hwaran Lee, Sangdoo Yun, and Seong Oh. 2024.
\newblock \href {https://doi.org/10.18653/v1/2024.acl-long.824} {Calibrating large language models using their generations only}.
\newblock In \emph{Proceedings of the 62nd Annual Meeting of the Association for Computational Linguistics (Volume 1: Long Papers)}, pages 15440--15459, Bangkok, Thailand. Association for Computational Linguistics.

\bibitem[{Wang et~al.(2024)Wang, Szarvas, Balazs, Danchenko, and Ernst}]{wang2024calibrating}
Cheng Wang, Gyuri Szarvas, Georges Balazs, Pavel Danchenko, and Patrick Ernst. 2024.
\newblock Calibrating verbalized probabilities for large language models.
\newblock \emph{arXiv preprint arXiv:2410.06707}.

\bibitem[{Wang et~al.(2020)Wang, Tu, Shi, and Liu}]{wang-etal-2020-inference}
Shuo Wang, Zhaopeng Tu, Shuming Shi, and Yang Liu. 2020.
\newblock \href {https://doi.org/10.18653/v1/2020.acl-main.278} {On the inference calibration of neural machine translation}.
\newblock In \emph{Proceedings of the 58th Annual Meeting of the Association for Computational Linguistics}, pages 3070--3079, Online. Association for Computational Linguistics.

\bibitem[{Wang and Holmes(2024)}]{wang2024subjective}
Ziyu Wang and Chris Holmes. 2024.
\newblock On subjective uncertainty quantification and calibration in natural language generation.
\newblock \emph{arXiv preprint arXiv:2406.05213}.

\bibitem[{Wiegreffe et~al.(2023)Wiegreffe, Finlayson, Tafjord, Clark, and Sabharwal}]{wiegreffe-etal-2023-increasing}
Sarah Wiegreffe, Matthew Finlayson, Oyvind Tafjord, Peter Clark, and Ashish Sabharwal. 2023.
\newblock \href {https://doi.org/10.18653/v1/2023.emnlp-main.522} {Increasing probability mass on answer choices does not always improve accuracy}.
\newblock In \emph{Proceedings of the 2023 Conference on Empirical Methods in Natural Language Processing}, pages 8392--8417, Singapore. Association for Computational Linguistics.

\bibitem[{Wu et~al.(2025)Wu, Lei, and Monz}]{wu2025calibrating}
Di~Wu, Yibin Lei, and Christof Monz. 2025.
\newblock Calibrating translation decoding with quality estimation on llms.
\newblock \emph{arXiv preprint arXiv:2504.19044}.

\bibitem[{Xiong et~al.(2023)Xiong, Hu, Lu, Li, Fu, He, and Hooi}]{xiong2023can}
Miao Xiong, Zhiyuan Hu, Xinyang Lu, Yifei Li, Jie Fu, Junxian He, and Bryan Hooi. 2023.
\newblock Can llms express their uncertainty? an empirical evaluation of confidence elicitation in llms.
\newblock \emph{arXiv preprint arXiv:2306.13063}.

\bibitem[{Yang et~al.(2024)Yang, Tsai, and Yamada}]{yang2024verbalized}
Daniel Yang, Yao-Hung~Hubert Tsai, and Makoto Yamada. 2024.
\newblock On verbalized confidence scores for llms.
\newblock \emph{arXiv preprint arXiv:2412.14737}.

\bibitem[{Yang et~al.(2023)Yang, Meng, Yan, and Zhou}]{yang-etal-2023-rethinking}
Zhen Yang, Fandong Meng, Yuanmeng Yan, and Jie Zhou. 2023.
\newblock \href {https://doi.org/10.18653/v1/2023.findings-acl.126} {Rethinking the word-level quality estimation for machine translation from human judgement}.
\newblock In \emph{Findings of the Association for Computational Linguistics: ACL 2023}, pages 2012--2025, Toronto, Canada. Association for Computational Linguistics.

\bibitem[{Zhang et~al.(2024)Zhang, Huang, Shi, Guo, Peng, Yan, Zhou, and Qiu}]{zhang-etal-2024-calibrating}
Mozhi Zhang, Mianqiu Huang, Rundong Shi, Linsen Guo, Chong Peng, Peng Yan, Yaqian Zhou, and Xipeng Qiu. 2024.
\newblock \href {https://doi.org/10.18653/v1/2024.emnlp-main.173} {Calibrating the confidence of large language models by eliciting fidelity}.
\newblock In \emph{Proceedings of the 2024 Conference on Empirical Methods in Natural Language Processing}, pages 2959--2979, Miami, Florida, USA. Association for Computational Linguistics.

\bibitem[{Zheng et~al.(2023)Zheng, Chiang, Sheng, Zhuang, Wu, Zhuang, Lin, Li, Li, Xing et~al.}]{zheng2023judging}
Lianmin Zheng, Wei-Lin Chiang, Ying Sheng, Siyuan Zhuang, Zhanghao Wu, Yonghao Zhuang, Zi~Lin, Zhuohan Li, Dacheng Li, Eric Xing, and 1 others. 2023.
\newblock Judging llm-as-a-judge with mt-bench and chatbot arena.
\newblock \emph{Advances in neural information processing systems}, 36:46595--46623.

\end{thebibliography}

\newpage
\appendix

\section{Prompts}
\label{sec:appenidx_prompts}

Table \ref{tab:all_prompts} shows all the prompts for the verbalized methods.

\section{Statistical Significance Results}
\label{app:significance_testing}
We report the full statistical significance results for both binary error detection and calibration. Following \citet{berg-kirkpatrick-etal-2012-empirical}, we use the bootstrap method, and collect 10000 values of test statistic for each comparison. To control family-wise error rate (FWER), we use the Holm-Bonferroni method. We choose $\alpha = 0.05$. We report the results for all 5 language directions for binary error detection in Tables \ref{tab:binary_error_statistical_significance_1} and \ref{tab:binary_error_statistical_significance_2}. Tables \ref{tab:calibration_statistical_significance_1} and \ref{tab:calibration_statistical_significance_2} show the results for calibration. Finally, Tables \ref{tab:correlation_statistical_significance_1} and \ref{tab:correlation_statistical_significance_2} present the results for the correlation of internal and verbalized methods.

\section{Verbalized $\Leftrightarrow$ Internal Extra Analysis}
\label{app:verbal_internal}

Although we do not see evidence of higher correlation between these two components, this does not mean they are any less correlated than (verbalized $\leftrightarrow$ ground-truth) and (internal $\leftrightarrow$ ground-truth). To be able to more directly compare them, we should at least use the same metric. For this, we binarize the internal methods to fit the results of the list method. Similarly to the alignments with the ground-truth, we use the dev sets to find binarization thresholds. Table \ref{tab:list_verbalized_internal_f1_test_results} shows the results for F1 on the negative labels. The much higher performance here is because the list method (and some other verbalized methods) has a high recall and lower precision (see Figure \ref{fig:open_vs_closed}). Therefore, both probability and entropy find that according to the dev set, the threshold that maximizes the F1 score for the desired tokens (i.e. tokens that are wrong and need to be changed, and we want to detect) is one which marks almost everything as wrong.  For that reason, directly comparing (internal $\leftrightarrow$ verbalized) with (internal $\leftrightarrow$ ground-truth) and (verbalized $\leftrightarrow$ ground-truth) is not very informative. We also report AUROC scores as a threshold-independent metric. Table \ref{tab:auroc_all_test_results} includes the AUROC scores for (internal $\leftrightarrow$ ground-truth) and (verbalized $\leftrightarrow$ ground-truth), and Table \ref{tab:list_verbalized_internal_auroc_test_results} reports the same for (internal $\leftrightarrow$ verbalized). We see that the alignments are generally on the same level, but (internal $\leftrightarrow$ verbalized) is less than most of the scores for (internal $\leftrightarrow$ ground-truth) and (verbalized $\leftrightarrow$ ground-truth) on average. Here is an example of the misalgnment between verbalized and internal signals:

\begin{itemize}
    \item \texttt{Source sentence: } Plougheth mine feeldes.
    \item \texttt{Model output: } \foreignlanguage{russian}{Мои поля орошаются} (the mistake is instead of plowing, the translation points to irrigation)
    \item \texttt{Gold annotations: } [1, 1, 1, 0, 0, 0, 0] (\foreignlanguage{russian}{орошаются} is wrong)

    \item \texttt{Model probabilities: } [0.086, 0.876, 0.797, 0.174, 0.93, 0.914, 0.894]

    \item \texttt{Verbalized confidence: }  ["a", "a", "a", "f", "f", "f", "f"], which we convert to: [1, 1, 1, 0, 0, 0, 0]  
\end{itemize}

Other than the first token which has a very low probability, we can see that model probabilities drop on the first token of the word \foreignlanguage{russian}{‘орошаются’}, but the rest of them are high probabilities. Spearman correlation coefficient between the verbalized numbers and the internal signals here is about -0.577.

\section{POS-specific Results}
\label{sec:app_pos_specific_charts}

\begin{figure*}[h]
    \centering
    \includegraphics[width=1\linewidth]{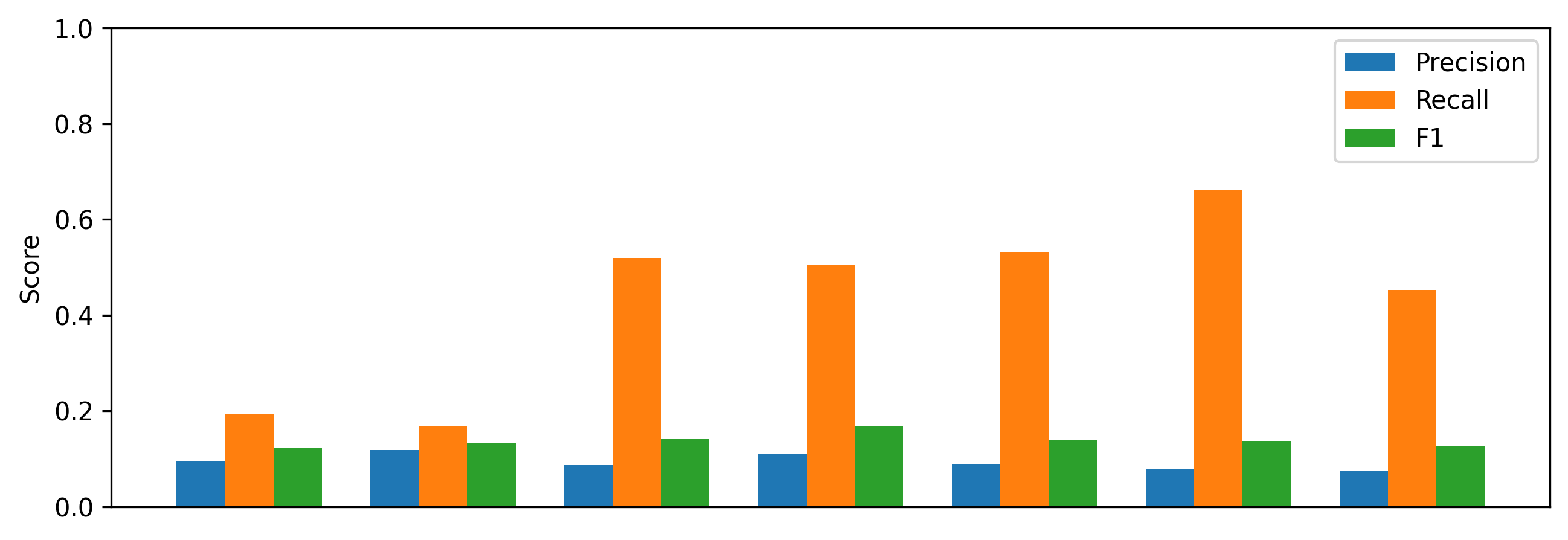} \\
    \includegraphics[width=1\linewidth]{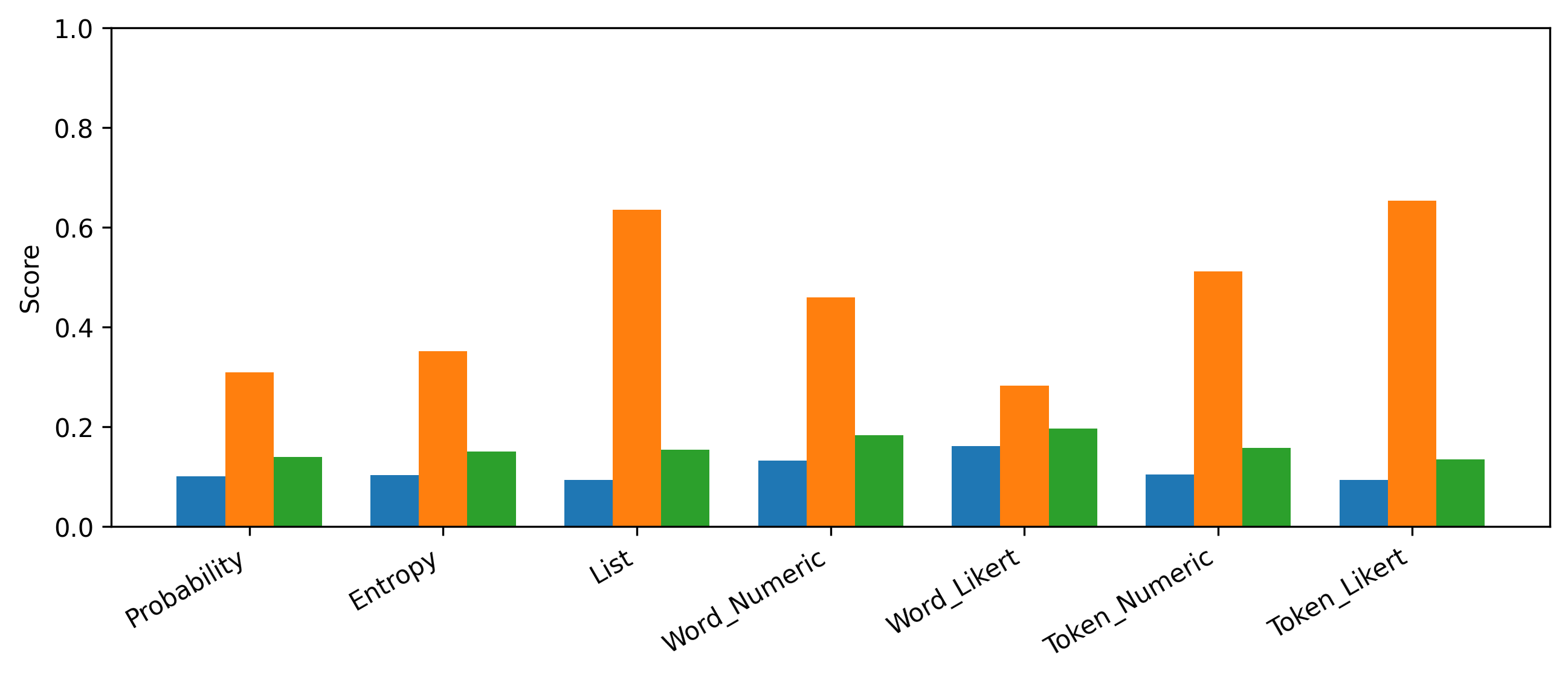}
    \caption{F1, precision, and recall scores for different methods averaged over all the languages on the test set for Aya23 (top) and Llama3-70B (bottom). These are for tokens that particularly have the \bl{ADJ} (adjective) label.}
    \label{fig:test-adj-charts}
\end{figure*}

\begin{figure*}[h]
    \centering
    \includegraphics[width=1\linewidth]{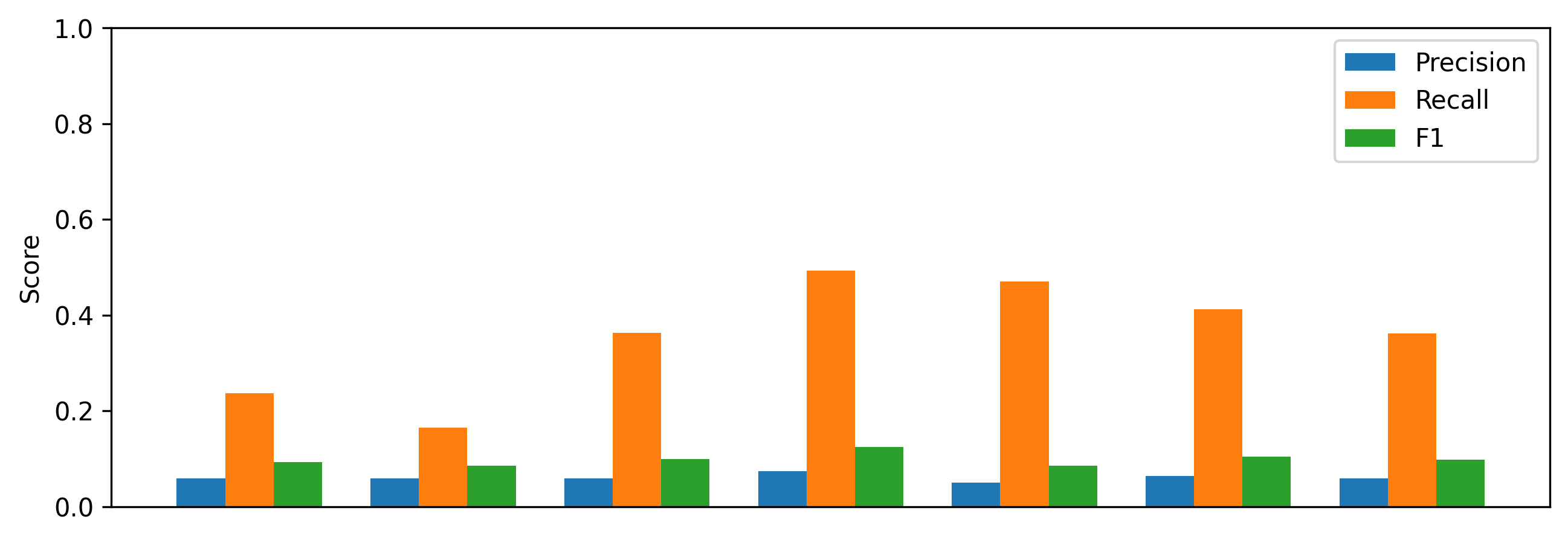} \\
    \includegraphics[width=1\linewidth]{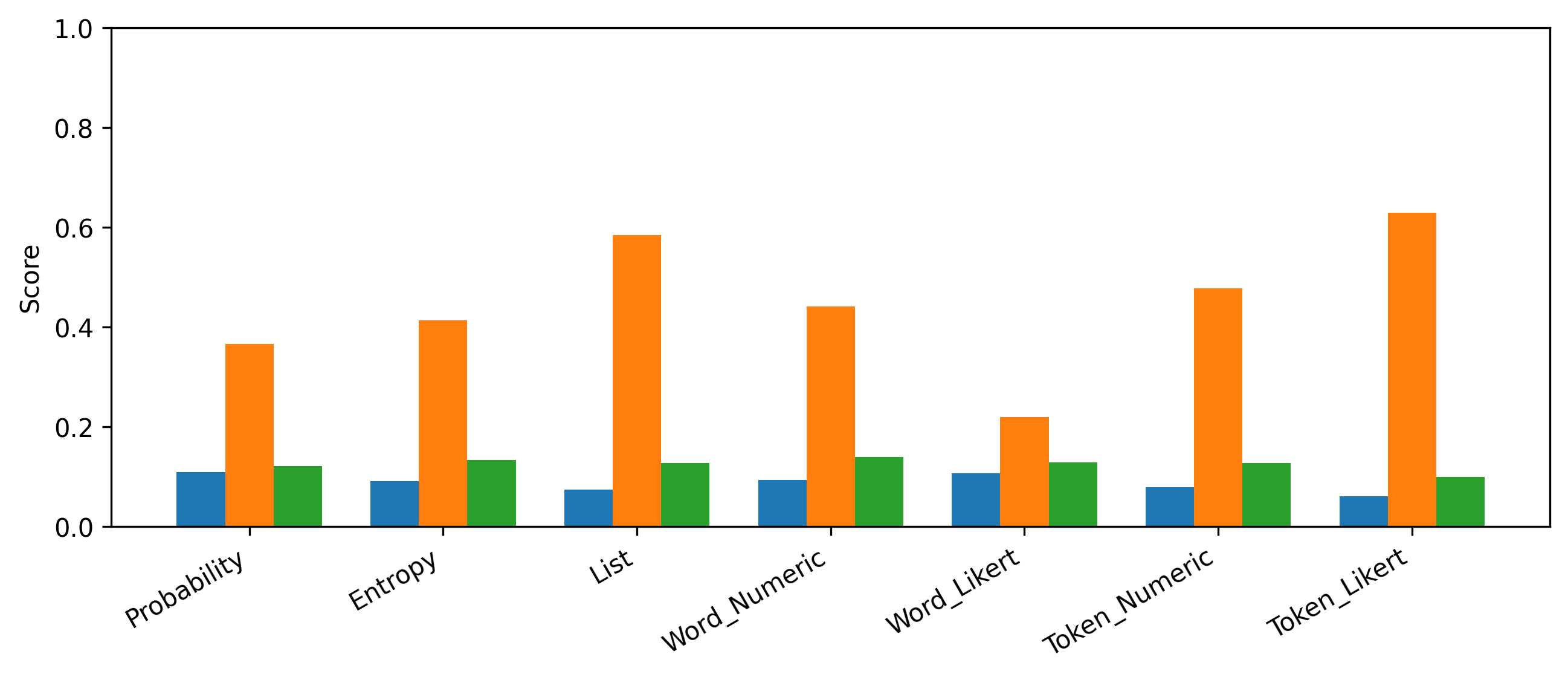}
    \caption{F1, precision, and recall scores for different methods averaged over all the languages on the test set for Aya23 (top) and Llama3-70B (bottom). These are for tokens that particularly have the \bl{ADV} (adverb) label.}
    \label{fig:test-adv-charts}
\end{figure*}

\begin{figure*}[h]
    \centering
    \includegraphics[width=1\linewidth]{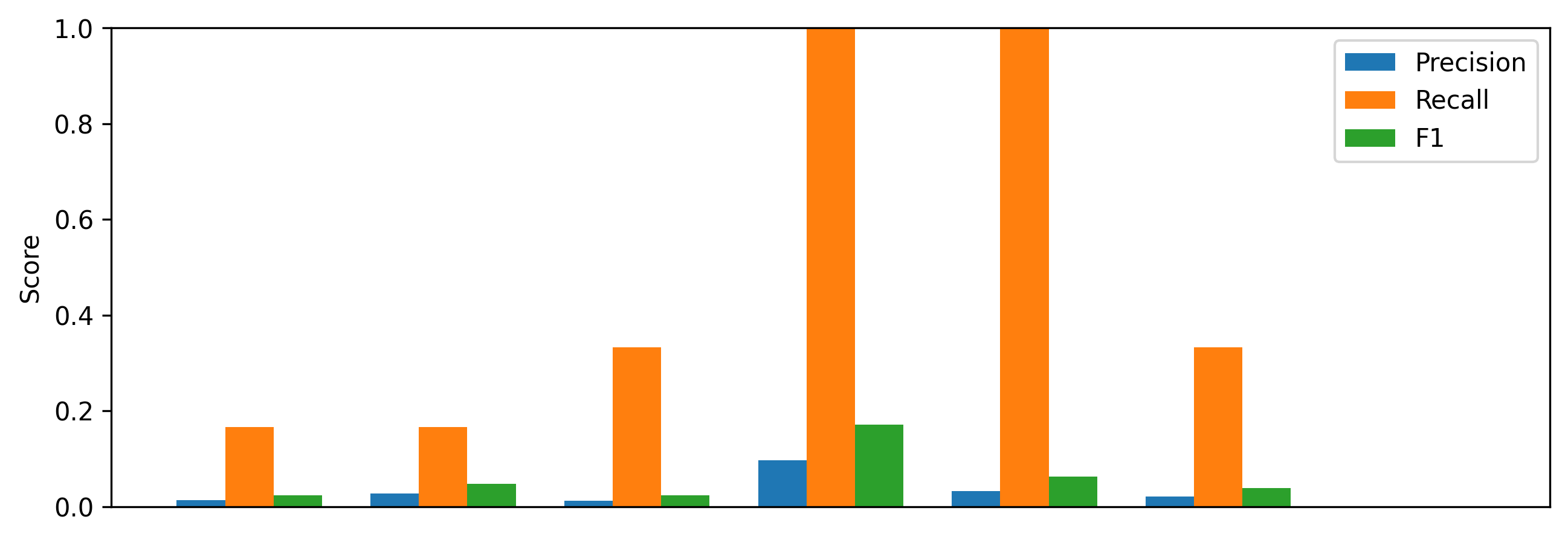} \\
    \includegraphics[width=1\linewidth]{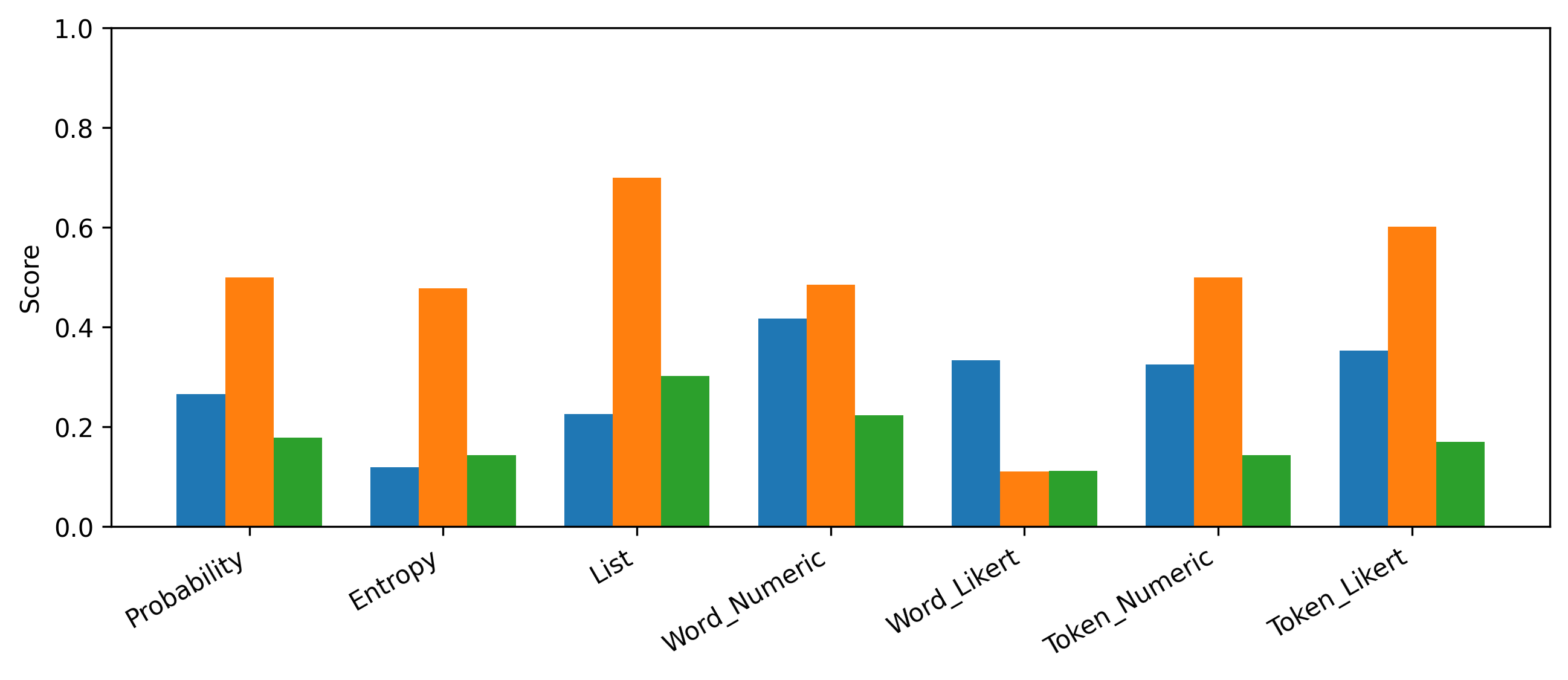}
    \caption{F1, precision, and recall scores for different methods averaged over all the languages on the test set for Aya23 (top) and Llama3-70B (bottom). These are for tokens that particularly have the \bl{INTJ} (interjection) label.}
    \label{fig:test-intj-charts}
\end{figure*}

\begin{figure*}[h]
    \centering
    \includegraphics[width=1\linewidth]{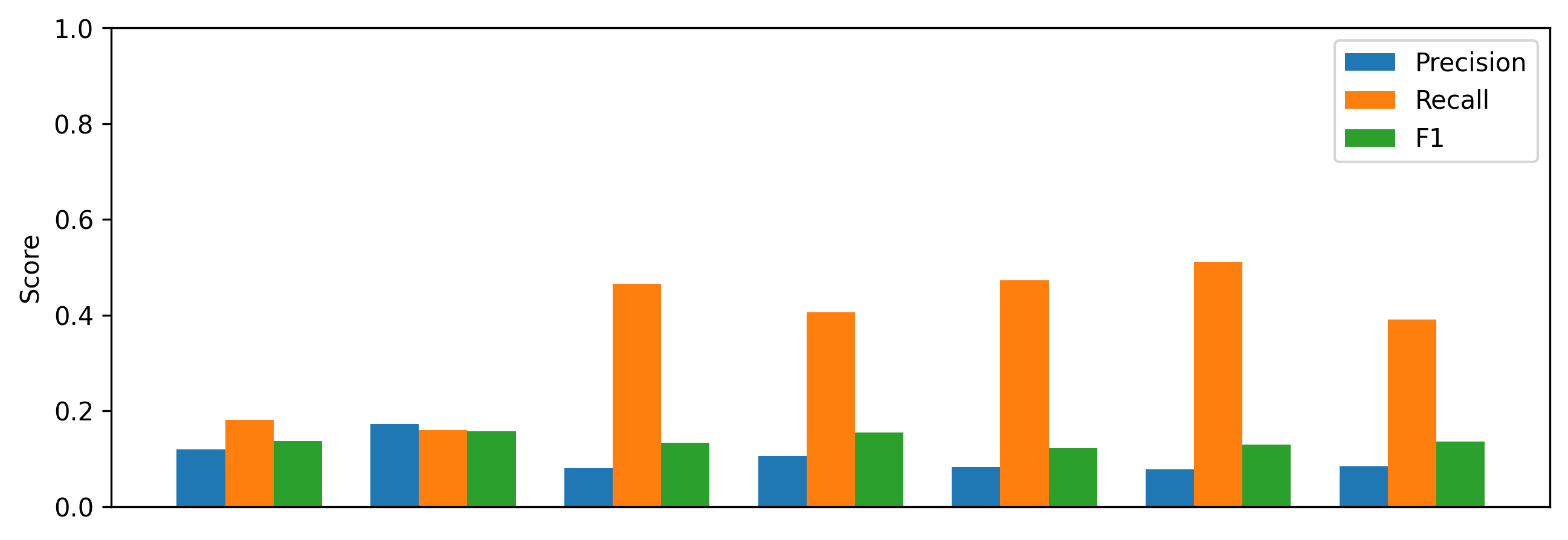} \\
    \includegraphics[width=1\linewidth]{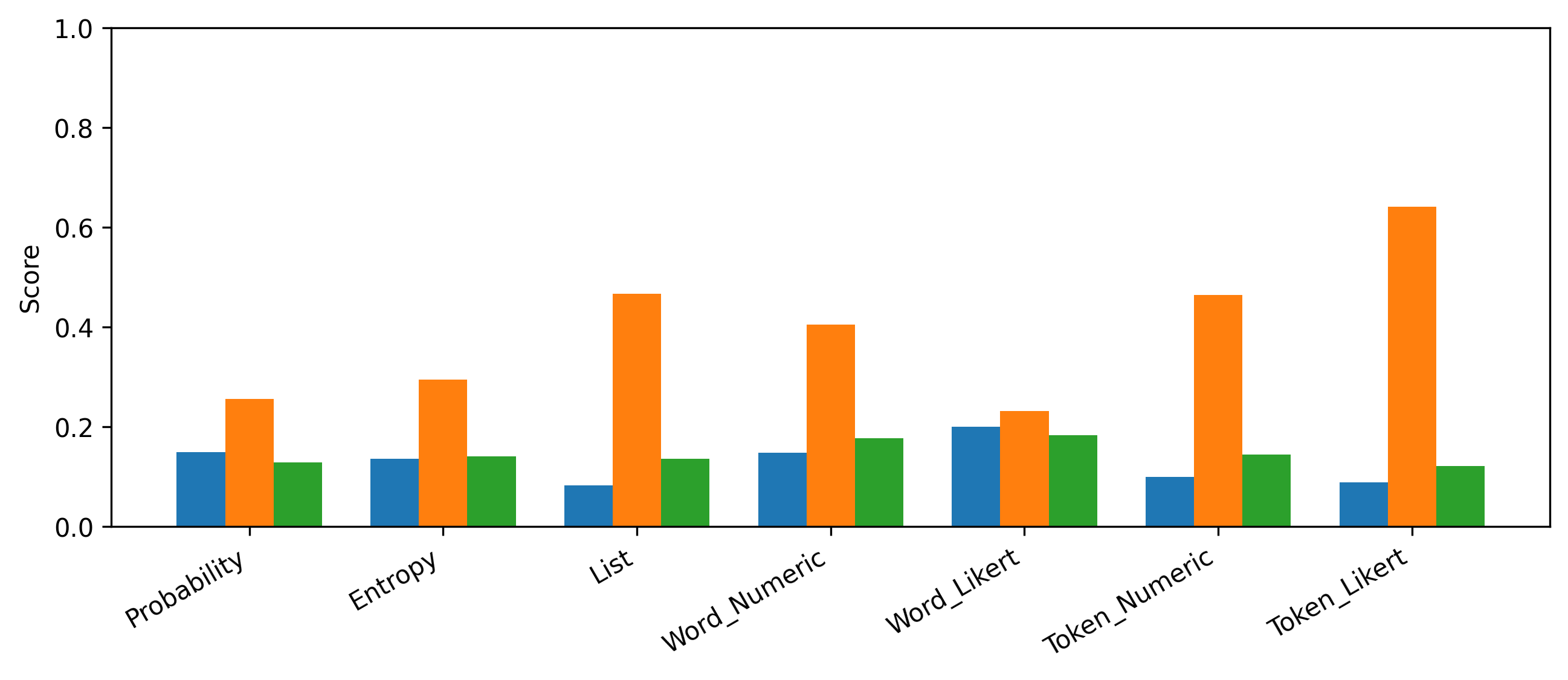}
    \caption{F1, precision, and recall scores for different methods averaged over all the languages on the test set for Aya23 (top) and Llama3-70B (bottom). These are for tokens that particularly have the \bl{NOUN} (noun) label.}
    \label{fig:test-noun-charts}
\end{figure*}

\begin{figure*}[h]
    \centering
    \includegraphics[width=1\linewidth]{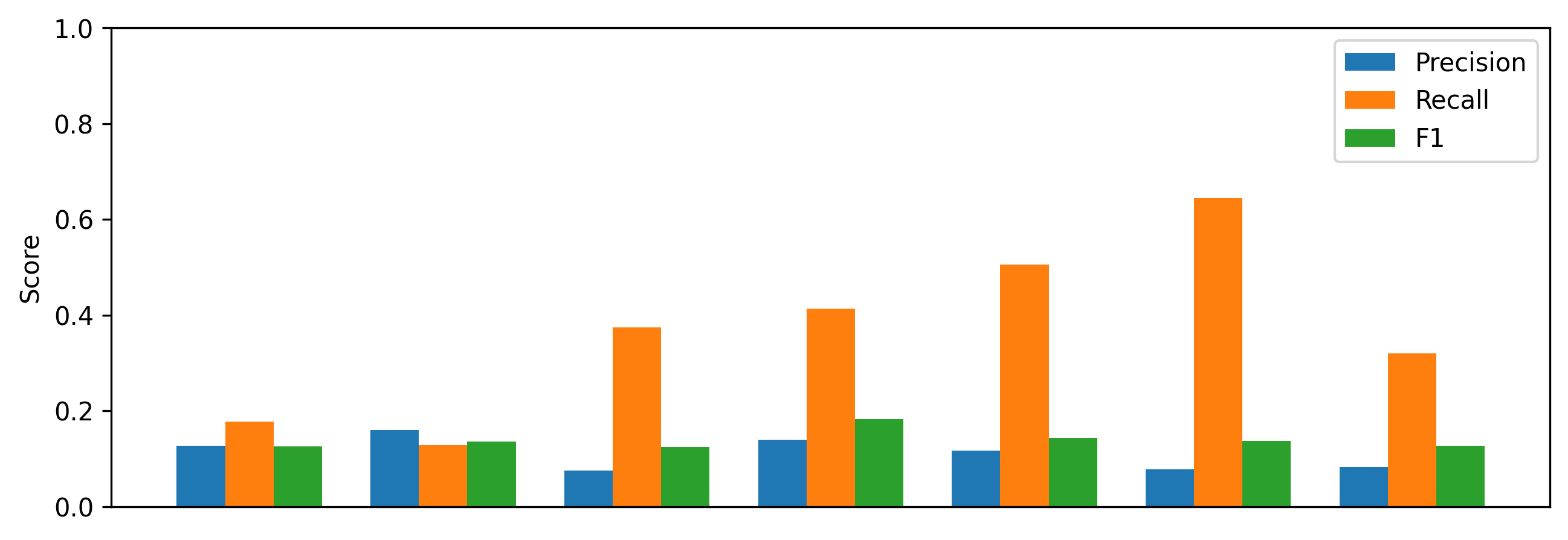} \\
    \includegraphics[width=1\linewidth]{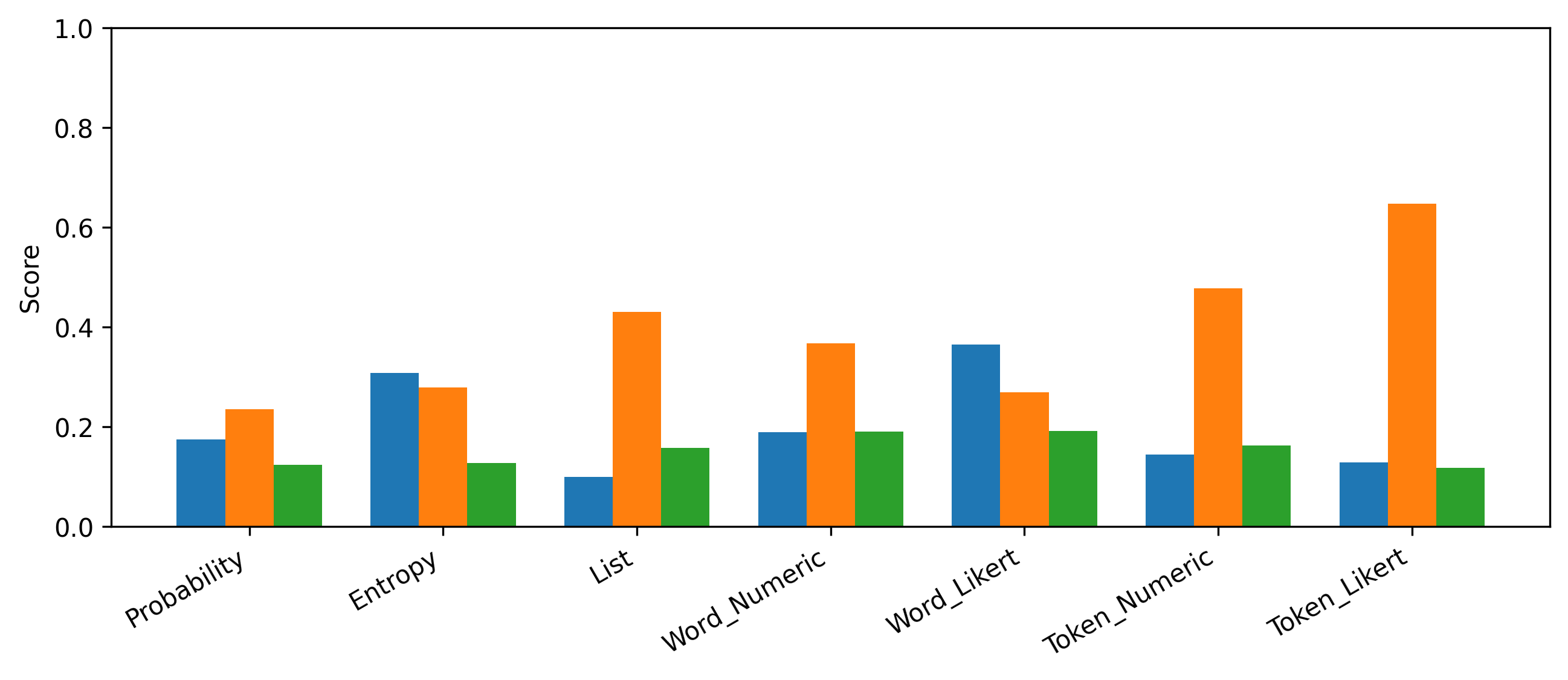}
    \caption{F1, precision, and recall scores for different methods averaged over all the languages on the test set for Aya23 (top) and Llama3-70B (bottom). These are for tokens that particularly have the \bl{PROPN} (proper noun) label.}
    \label{fig:test-propn-charts}
\end{figure*}

\begin{figure*}[h]
    \centering
    \includegraphics[width=1\linewidth]{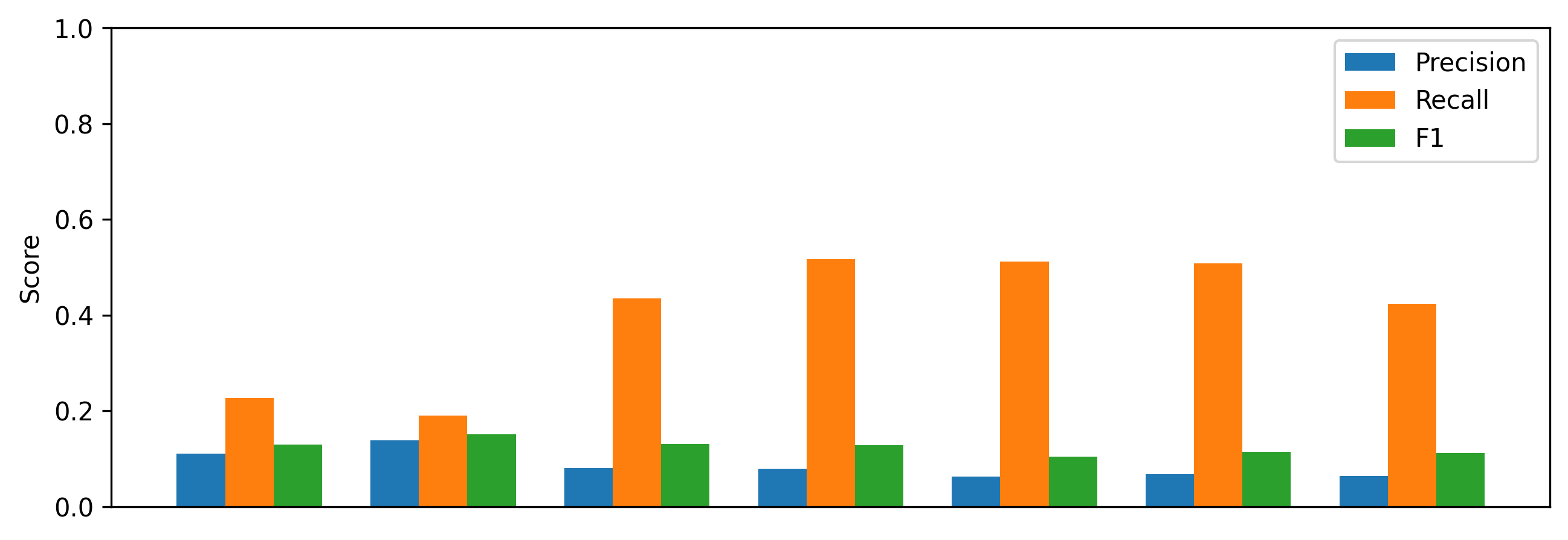} \\
    \includegraphics[width=1\linewidth]{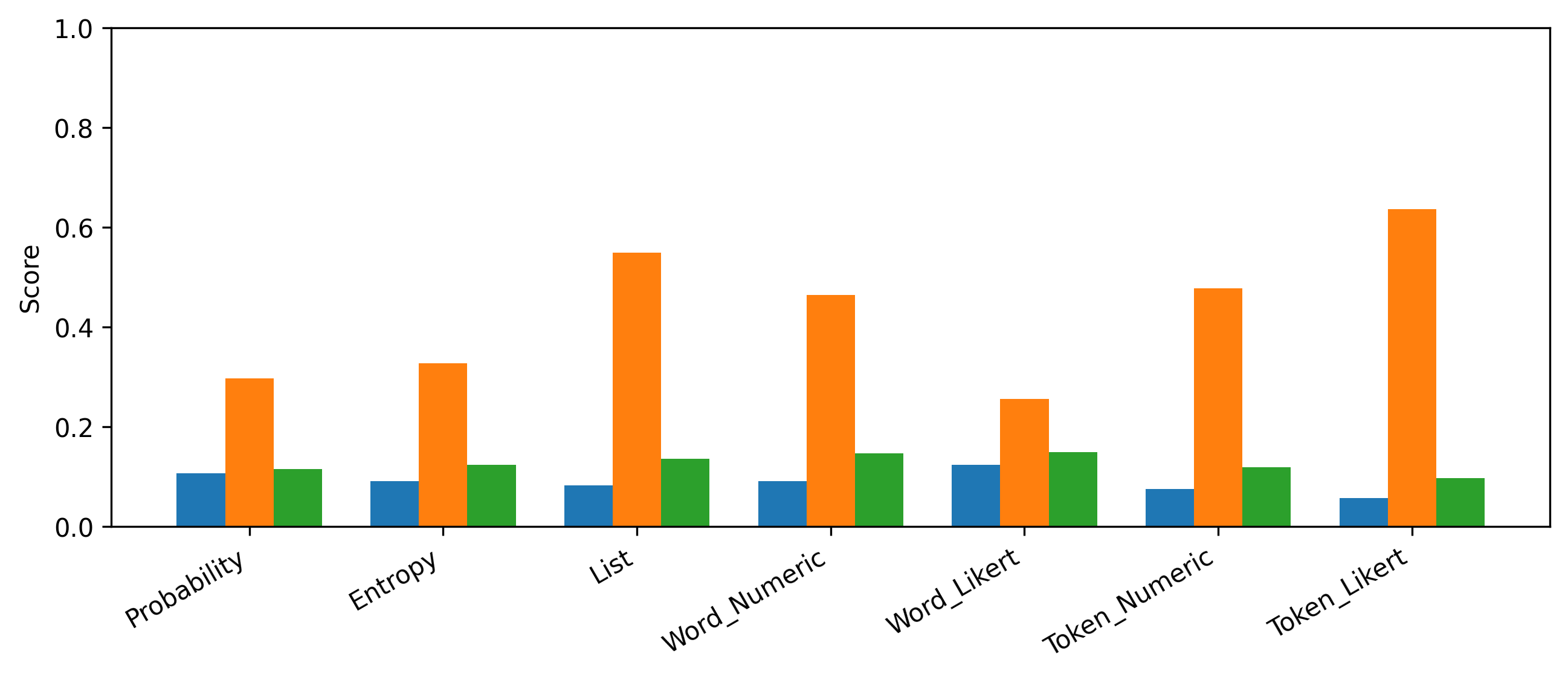}
    \caption{F1, precision, and recall scores for different methods averaged over all the languages on the test set for Aya23 (top) and Llama3-70B (bottom). These are for tokens that particularly have the \bl{VERB} (verb) label.}
    \label{fig:test-verb-charts}
\end{figure*}

Figures \ref{fig:test-adj-charts} through \ref{fig:test-verb-charts} showcase the results of all the methods' performances for each ``open'' syntactic role as labeled by Stanza: adjectives, adverbs, interjections, nouns, proper nouns, and verbs. \\
We can see some interesting patterns in these charts. For adjectives, Word\_Likert and Word\_Numeric have the best performances on average. 
For interjections, Word\_Numeric  is \textit{by far} the best performing method for Aya23, while only falling short of the List method for Llama3-70B. \\
For nouns, Word\_Numeric and Word\_Likert lead for Llama3-70B. For Aya23, Entropy replaces Word\_Likert as the second best method. \\
Verbalized methods also dominate the performances for proper nouns for Llama3-70B. For Aya23, Word\_Numeric still outperforms other methods. \\
For verbs, Entropy outperforms other methods for Aya23. For Llama3-70B it lags behind List, Word\_Numeric, and Word\_Likert.

\section{Other Results}
\label{extra_results}

\subsection{Dev Set-Independent Binarization}

Table \ref{tab:f1_optimal_all_test_results} shows the results for binary error detection if we artificially binarize the scores for the listed methods optimally on the test set. This serves as an upper bound on performance of these methods, and is consistent with the analysis of \citet{sarti-etal-2025-unsupervised}. Note that Random and List methods do not need binarization and report the same performances as Table \ref{tab:f1_threshold_all_test_results}. \\
Overall, the trends are consistent. Word\_Numeric continues to perform well on average, with Entropy showing similar performance. Entropy performs best for Aya23, consistent with the findings of \citet{sarti-etal-2025-unsupervised}. In contrast, three verbalized methods outperform Entropy in the Llama3-70B experiments.

\begin{table*}[t]
\footnotesize
\centering
\begin{tabular}{l|cc|cc|cc|cc|cc|cc}
\toprule[0.5pt]
\multicolumn{1}{c}{\multirow{2}{*}{Method}} & \multicolumn{2}{c}{\textbf{En $\rightarrow$ Cs}} & \multicolumn{2}{c}{\textbf{En $\rightarrow$ Hi}} & \multicolumn{2}{c}{\textbf{En $\rightarrow$ Ja}} & \multicolumn{2}{c}{\textbf{En $\rightarrow$ Ru}} & \multicolumn{2}{c}{\textbf{En $\rightarrow$ Zh}} & \multicolumn{2}{c}{\textbf{Average}}\\
\cmidrule(lr){2-3}
\cmidrule(lr){4-5}
\cmidrule(lr){6-7}
\cmidrule(lr){8-9}
\cmidrule(lr){10-11}
\cmidrule(lr){12-13}

& Aya  & Llama  & Aya  & Llama  & Aya  & Llama  & Aya  & Llama  & Aya  & Llama & Aya  & Llama \\
\midrule
Random &        0.07&0.09&0.03&0.03&0.01&0.01&0.08&0.11&0.03&0.04&0.04&0.06     \\
\midrule
Probability &0.17&0.20&0.11&0.09&0.07&0.08&0.19&0.20&0.11&0.11&0.13&0.14     \\

Entropy&\bl{0.19}&0.20&0.11&0.10&\bl{0.10}&0.09&0.20&0.20&\bl{0.13}&0.13&0.15&0.14 \\
\midrule
List&0.15&0.20&\bl{0.13}&0.08&0.05&0.08&0.18&0.22&0.09&0.11&0.12&0.14 \\
Word\_Numeric&\bl{0.19}&\bl{0.22}&0.12&\bl{0.17}&0.06&0.13&\bl{0.21}&0.22&0.11&0.14&0.14&0.18 \\
Word\_Likert&0.15&\bl{0.22}&0.11&0.15&0.05&\bl{0.15}&0.15&\bl{0.23}&0.08&\bl{0.16}&0.11&0.18 \\
Token\_Numeric&0.16&0.20&0.09&0.10&0.06&0.10&0.19&0.21&0.11&0.14&0.12&0.15 \\
Token\_Likert&0.16&0.16&0.11&0.08&0.05&0.10&0.19&0.16&0.10&0.14&0.12&0.13 \\
\bottomrule[0.5pt]
\end{tabular}
\caption{Results for the binary error detection task on the test set. All the numbers report the F1 score of negative labels (higher is better), where continuous methods are \bl{optimally binarized}. Best performing results are bolded in each column. }
\label{tab:f1_optimal_all_test_results}
\end{table*}

\subsection{Bundling}

\label{budling_definition}

For the methods that do not operate on the word level, we want to measure how the methods ``bundle'' their allegedly wrong tokens (according to their prediction) together. For this, we consider \textit{all} the words that have \textit{any} wrong tokens, and calculate what ratio of all the tokens of these words are predicted as wrong. For example, supposing we have only one word and it has 4 tokens and the model predictions are [0, 0, 0, 1], then the bundling ratio would be (number of wrong tokens / summation of the length of all the words with any wrong tokens) = (3 / 4) = 0.75. Table \ref{tab:bundling_ratios} shows the bundling ratio of different models as aggregated across different languages on the test set. We see that for both models, 
 verbalized methods have higher concentration of wrong tokens in the same words than internal methods. This is closer to the way humans annotated the dataset, since they specified errors on the word level.

\subsection{Prompt Costs}
\label{prompt_cost_report}
We approximate the cost of each verbalized method by reporting the average number of tokens for each prompt. Table \ref{tab:prompt_costs} shows the results.

\subsection{Computing Infrastructure}

We used a mixture of NVIDIA 80GB H100 SXM GPUs and NVIDIA A100 Tensor Core GPUs in our experiments. Aya23, at 35B parameters, could be prompted by a single H100 but Llama3-70B needs more GPUs.

\begin{table*}
    \centering
    \small
    \begin{tabular}{lp{12.5cm}}
        \toprule
        \textbf{Method} & \textbf{Template} \\
        \midrule
         Shared prefix (all methods) & \texttt{An LLM was asked to translate from \{SOURCE\_LANGUAGE\} to \{TARGET\_LANGUAGE\}, given the following prompt: \textbackslash n '''\{ORIGINAL\_PROMPT\}'''\textbackslash n\textbackslash nThe LLM produced the following translation, Please identify any part that might have issues in this translation: \textbackslash n'''\{TRANSLATION\}'''\textbackslash n\textbackslash n
         }
         \\
         \midrule
         \midrule
         
         List & \texttt{Your task is to identify every span in the translation that does not perfectly align with the source text. Scrutinize the translation for ANY deviation. \textbackslash nEven if a phrase is generally understandable, you should mark it if it could be rendered more precisely or naturally. Example output format: [`span1', `span2', `span3'] \textbackslash n\textbackslash nTranslation is: \{TRANSLATION\}\textbackslash n Please make sure the strings are exact substrings of the translation. Do not give me any explanations, your output should be only a list:\textbackslash n\textbackslash n
        }  \\
         \midrule
         Word\_Numeric& \texttt{Your task is to identify every span in the translation that does not perfectly align with the source text. Scrutinize the translation for any deviation, including not just direct mistranslations but also subtle shifts in meaning, tone, style, and register. \textbackslash nEven if a phrase is generally understandable, you should mark it if it could be rendered more precisely or naturally.\textbackslash nFor each word, we want you to return a confidence score marking your certainty in it being correct - not having any of the issues we mentioned. If uncertain about a word, the score should be low (less than or around 0.2), but if certain it should be high (0.8 or above). All the words of the translation are: \{ALL\_WORDS\}. How confident are you about the \{PLACEMENT\} word, \{WORD\_OF\_INTEREST\}, being correct in the translation? \textbackslash nONLY return a confidence score from 0.0 (uncertain) to 1.0 (certain) with no extra explanations or words, your output should simply be a single number: \textbackslash n}\\  
         \midrule
         Word\_Likert& \texttt{We want to identify the text fragments that contain a translation error, regardless of whether the error is major (like altered meaning, reduced readability or usability) or minor (like unnatural style, poor grammar or lexical choice). \textbackslash nIf something is missing in the translation, we disregard it for now. We will go through each word, one by one, to get a confidence score from you marking your certainty about it being correct. Use the following scale: a. very certain \textbackslash n b. fairly certain \textbackslash n c. moderately certain \textbackslash n d. somewhat certain \textbackslash n e. not certain \textbackslash n f. very uncertain \textbackslash n \textbackslash nAll the words of the translation are: \{ALL\_WORDS\}. How confident are you about the \{PLACEMENT\} word, \{WORD\_OF\_INTEREST\}, being correct in the translation? \textbackslash nONLY return a letter indicating your confidence, from a (for absolute certainty) to f (very uncertain) with no extra explanations or words, your output should simply be a single letter: \textbackslash n} \\
         \midrule
         Token\_Numeric& \texttt{Your task is to identify every span in the translation that does not perfectly align with the source text. Scrutinize the translation for any deviation, including not just direct mistranslations but also subtle shifts in meaning, tone, style, and register. \textbackslash nEven if a phrase is generally understandable, you should mark it if it could be rendered more precisely or naturally.\textbackslash nFor each token, we want you to return a confidence score marking your certainty in it being correct - not having any of the issues we mentioned. If uncertain about a token, the score should be low (less than or around 0.2), but if certain it should be high (0.8 or above). \textbackslash nAll the tokens of the translation are: \{ALL\_TOKENS\}. How confident are you about the \{PLACEMENT\} token, \{TOKEN\_OF\_INTEREST\}, being correct in the translation? ONLY return a confidence score from 0.0 (uncertain) to 1.0 (certain) with no extra explanations or words, your output should simply be a single number: \textbackslash n} \\
         \midrule
         Token\_Likert& \texttt{For every token, give me a score as to how sure you are about it being correct, doing it one at a time. Use the following scale: a. very certain \textbackslash n b. fairly certain \textbackslash n c. moderately certain \textbackslash n d. somewhat certain \textbackslash n e. not certain \textbackslash n f. very uncertain \textbackslash n \textbackslash nAll the tokens of the translation are: \{ALL\_TOKENS\}. How confident are you about the \{PLACEMENT\} token, \{TOKEN\_OF\_INTEREST\}, being correct in the translation? ONLY return a letter indicating your confidence, from a (for absolute certainty) to f (very uncertain) with no extra explanations or words, your output should simply be a single letter: \textbackslash n } \\
         \bottomrule
    \end{tabular}
    \caption{Prompt templates for each verbalized method.}
    \label{tab:all_prompts}
\end{table*}

\begin{table*}[]
\footnotesize
\centering
\begin{tabular}{l|c|c|c|c|c|c}
\toprule[0.5pt]
\multicolumn{1}{c}{\multirow{2}{*}{Method}} & \multicolumn{2}{c}{\textbf{En $\rightarrow$ Cs}} & \multicolumn{2}{c}{\textbf{En $\rightarrow$ Hi}} & \multicolumn{2}{c}{\textbf{En $\rightarrow$ Ja}} \\
\cmidrule(lr){2-3}
\cmidrule(lr){4-5}
\cmidrule(lr){6-7}

& Aya  & Llama  & Aya  & Llama  & Aya  & Llama\\
\midrule

Probability (A)&B&E&B C D E&D E&B C D F G&C D E F G\\
Entropy (B)&---&E&C&D E&---&C D E F G\\
\midrule
List (C)&A B D&E&---&D E F&B D F&D F\\
Word\_Numeric (D)&---&E&C&E&---&---\\
Word\_Likert (E)&A B C D F G&---&C&---&B D&D\\
Token\_Numeric (F)&B D&E&B C D E&D E&B&D\\
Token\_Likert (G)&A B D F&A B C D E F&B C D E&D E F&B&D\\
\bottomrule[0.5pt]
\end{tabular}
\caption{Statistical significance results for binary error detection as measured in F1 on the wrong (negative) tokens. These are the results for English $\rightarrow$ Czech, Hindi, and Japanese. We employ the bootstrap method. For each method, we report the methods that are \textit{significantly better} in the column (language, model pair).}

\label{tab:binary_error_statistical_significance_1}
\end{table*}

\begin{table*}[]
\footnotesize
\centering
\begin{tabular}{l|c|c|c|c}
\toprule[0.5pt]
\multicolumn{1}{c}{\multirow{2}{*}{Method}} & \multicolumn{2}{c}{\textbf{En $\rightarrow$ Ru}} & \multicolumn{2}{c}{\textbf{En $\rightarrow$ Zh}} \\
\cmidrule(lr){2-3}
\cmidrule(lr){4-5}

& Aya  & Llama  & Aya  & Llama \\
\midrule

Probability (A)&D&C D E F&---&B E\\
Entropy (B)&---&A C D E F&A&---\\
\midrule
List (C)&D&---&A&E\\
Word\_Numeric (D)&---&C E&A G&E\\
Word\_Likert (E)&A B C D F G&---&A C G&---\\
Token\_Numeric (F)&D&C E&A G&E\\
Token\_Likert (G)&D&A B C D E F&---&B C D E F\\
\bottomrule[0.5pt]
\end{tabular}
\caption{Statistical significance results for binary error detection as measured in F1 on the wrong (negative) tokens. These are the results for English $\rightarrow$ Russian and Chinese. We employ the bootstrap method. For each method, we report the methods that are \textit{significantly better} in the column (language, model pair).}

\label{tab:binary_error_statistical_significance_2}
\end{table*}

\begin{table*}[]
\footnotesize
\centering
\begin{tabular}{l|c|c|c|c|c|c}
\toprule[0.5pt]
\multicolumn{1}{c}{\multirow{2}{*}{Method}} & \multicolumn{2}{c}{\textbf{En $\rightarrow$ Cs}} & \multicolumn{2}{c}{\textbf{En $\rightarrow$ Hi}} & \multicolumn{2}{c}{\textbf{En $\rightarrow$ Ja}} \\
\cmidrule(lr){2-3}
\cmidrule(lr){4-5}
\cmidrule(lr){6-7}

& Aya  & Llama  & Aya  & Llama  & Aya  & Llama\\
\midrule
Probability (A)&B E&B&B&B G&B E G&B E G\\
Entropy (B)&---&---&---&---&---&---\\
List (C)&A B D E G&A B D E F G&A B D E G&A B D E F G&A B E G&A B D E F G\\
Word\_Numeric (D)&A B E&A B E G&A B E G&A B E F G&A B C E G&A B E G\\
Word\_Likert (E)&B&A B&A B&A B G&---&B G\\
Token\_Numeric (F)&A B C D E G&A B E G&A B C D E G&A B E G&A B C D E G&A B E G\\
Token\_Likert (G)&A B D E&A B&A B E&B&B E&B\\
\bottomrule[0.5pt]
\end{tabular}
\caption{Statistical significance results for calibration as measured in ECE. These are the results for English $\rightarrow$ Czech, Hindi, and Japanese. We employ the bootstrap method. For each method, we report the methods that are \textit{significantly better} in the column (language, model pair).}

\label{tab:calibration_statistical_significance_1}
\end{table*}

\begin{table*}[]
\footnotesize
\centering
\begin{tabular}{l|c|c|c|c}
\toprule[0.5pt]
\multicolumn{1}{c}{\multirow{2}{*}{Method}} & \multicolumn{2}{c}{\textbf{En $\rightarrow$ Ru}} & \multicolumn{2}{c}{\textbf{En $\rightarrow$ Zh}}\\
\cmidrule(lr){2-3}
\cmidrule(lr){4-5}

& Aya  & Llama  & Aya  & Llama\\
\midrule

Probability (A)&B&B&B E&B E\\
Entropy (B)&---&---&---&---\\
List (C)&A B D E G&A B D E F G&A B D E&A B D E F G\\
Word\_Numeric (D)&A B E&A B E F G&A B E&A B E G\\
Word\_Likert (E)&B&A B G&B&B\\
Token\_Numeric (F)&A B C D E G&A B E G&A B C D E G&A B D E G\\
Token\_Likert (G)&A B D E&A B&A B C D E&A B E\\
\bottomrule[0.5pt]
\end{tabular}
\caption{Statistical significance results for calibration as measured in ECE. These are the results for English $\rightarrow$ Russian and Chinese. We employ the bootstrap method. For each method, we report the methods that are \textit{significantly better} in the column (language, model pair).}

\label{tab:calibration_statistical_significance_2}
\end{table*}

\begin{table*}
\footnotesize
\centering
{
\setlength{\tabcolsep}{4.5pt}
\begin{tabular}{cl|c|c|c|c|c|c|}
\toprule[0.5pt]
& \multicolumn{1}{c}{\multirow{2}{*}{Method}} & \multicolumn{2}{c}{\textbf{En $\rightarrow$ Cs}} & \multicolumn{2}{c}{\textbf{En $\rightarrow$ Hi}} & \multicolumn{2}{c}{\textbf{En $\rightarrow$ Ja}} \\
\cmidrule(lr){3-4}
\cmidrule(lr){5-6}
\cmidrule(lr){7-8}

& & Aya  & Llama  & Aya  & Llama  & Aya  & Llama \\

\multirow{4}{*}{\rotatebox{90}{Probability}}&List (C)&D&---&D E&D E&D&D E\\
&Word\_Numeric (D)&---&---&---&---&---&---\\
&Word\_Likert (E)&D&C D&D&D&C D&D\\
&Token\_Numeric (F)&C D E G&C D E&C D E G&C D E&C D E G&C D E\\
&Token\_Likert (G)&C D E&C D E F&C D E&C D E F&C D E&C D E F\\
\midrule
\multirow{4}{*}{\rotatebox{90}{Entropy}}
&List (C)&D&---&D E&D E&D&D E\\
&Word\_Numeric (D)&---&---&---&---&---&---\\
&Word\_Likert (E)&D&C D&D&D&C D&D\\
&Token\_Numeric (F)&C D E G&C D E&C D E G&C D E&C D E G&C D E\\
&Token\_Likert (G)&C D E&C D E F&C D E&C D E F&C D E&C D E F\\

\bottomrule[0.5pt]
\end{tabular}
}
\caption{Statistical significance results for correlation of internal and verbalized methods as measured in Spearman's coefficient. These are the results for English $\rightarrow$ Czech, Hindi, and Japanese. We employ the bootstrap method. For each method, we report the methods that are \textit{significantly better} in the column (language, model pair).}
\label{tab:correlation_statistical_significance_1}
\end{table*}

\begin{table*}
\footnotesize
\centering
{
\setlength{\tabcolsep}{4.5pt}
\begin{tabular}{cl|c|c|c|c|}
\toprule[0.5pt]
& \multicolumn{1}{c}{\multirow{2}{*}{Method}} & \multicolumn{2}{c}{\textbf{En $\rightarrow$ Ru}} & \multicolumn{2}{c}{\textbf{En $\rightarrow$ Zh}} \\
\cmidrule(lr){3-4}
\cmidrule(lr){5-6}

& & Aya  & Llama  & Aya  & Llama \\

\multirow{4}{*}{\rotatebox{90}{Probability}}&List (C)&D&D&D F&D E\\
&Word\_Numeric (D)&---&---&---&---\\
&Word\_Likert (E)&D&D&D F&D\\
&Token\_Numeric (F)&C D E G&C D&D&C D E\\
&Token\_Likert (G)&C D E&C D E F&D E F&C D E F\\
\midrule
\multirow{4}{*}{\rotatebox{90}{Entropy}}
&List (C)&D&D&D E F&D E\\
&Word\_Numeric (D)&---&---&---&---\\
&Word\_Likert (E)&D&D&D F&D\\
&Token\_Numeric (F)&C D E G&C D&D&C D E\\
&Token\_Likert (G)&C D E&C D E F&D E F&C D E F\\

\bottomrule[0.5pt]
\end{tabular}
}
\caption{Statistical significance results for correlation of internal and verbalized methods as measured in Spearman's coefficient. These are the results for English $\rightarrow$ Russian and Chinese. We employ the bootstrap method. For each method, we report the methods that are \textit{significantly better} in the column (language, model pair).}
\label{tab:correlation_statistical_significance_2}
\end{table*}

\begin{table*}
\footnotesize
\centering
\begin{tabular}{l|c|c}
\toprule[0.5pt]
\multicolumn{1}{c}{\bl{Method}} & \multicolumn{1}{c}{\textbf{Aya23}} & \multicolumn{1}{c}{\textbf{Llama3-70B}} \\
\midrule
Probability & 0.29 & 0.40\\
Entropy & 0.21 & 0.39\\
\midrule
Token\_Likert & 0.43& 0.77\\
Token\_Numeric & 0.56 & 0.47\\
\bottomrule[0.5pt]
\end{tabular}
\caption{The ``bundling'' ratio of different models, as describe in \ref{budling_definition}. We aggregate the tokens of the test sets of all different languages.}
\label{tab:bundling_ratios}
\end{table*}

\begin{table*}
\footnotesize
\centering
\setlength{\tabcolsep}{3pt}
\begin{tabular}{l|c|c|c|c|c|c|c|c|c|c}
\toprule[0.5pt]
\multicolumn{1}{c}{\multirow{2}{*}{Method}} & \multicolumn{2}{c}{\textbf{En $\rightarrow$ Cs}} & \multicolumn{2}{c}{\textbf{En $\rightarrow$ Hi}} & \multicolumn{2}{c}{\textbf{En $\rightarrow$ Ja}} & \multicolumn{2}{c}{\textbf{En $\rightarrow$ Ru}} & \multicolumn{2}{c}{\textbf{En $\rightarrow$ Zh}}\\
\cmidrule(lr){2-3}
\cmidrule(lr){4-5}
\cmidrule(lr){6-7}
\cmidrule(lr){8-9}
\cmidrule(lr){10-11}

& Aya  & Llama  & Aya  & Llama  & Aya  & Llama  & Aya  & Llama  & Aya  & Llama \\

List & 130.75 	&134.65 	& 209.26 	& 177.96 	& 127.14 	& 136.52 	& 128.77 	& 138.87 	& 117.13 	& 121.16
\\

Word\_Numeric 	& 6775.31 	& 6879.0 	& 12445.33 	& 10858.25 	& 12060.62 	& 12645.24 	& 6878.98 	& 7142.67 	& 8194.54 	& 8199.34 \\

Word\_Likert &	6032.8 &	6270.99 &	11464.89 &	10066.85 	& 10714.68 	& 11537.73 	& 6117.6 	& 6524.16 	& 7231.16 	& 7423.21 \\

Token\_Numeric 	& 11993.59 	& 13369.72 	&38610.03 	&26910.33 	&11066.73 	&13876.36 	&11482.26 	&14522.99 	&8633.06 	& 9918.55 \\

Token\_Likert 	& 7213.97 	& 8380.3 	& 27549.32 	&18629.04 	& 6575.76 	& 8744.94 	& 6860.88 	& 9213.11 	& 4942.4 	& 5954.45 \\
\end{tabular}
\caption{Average number of tokens for the prompt of each verbalized method. }
\label{tab:prompt_costs}
\end{table*}

\begin{table*}
\footnotesize
\centering
\begin{tabular}{l|cc|cc|cc|cc|cc|cc}
\toprule[0.5pt]
\multicolumn{1}{c}{\multirow{2}{*}{Method}} & \multicolumn{2}{c}{\textbf{En $\rightarrow$ Cs}} & \multicolumn{2}{c}{\textbf{En $\rightarrow$ Hi}} & \multicolumn{2}{c}{\textbf{En $\rightarrow$ Ja}} & \multicolumn{2}{c}{\textbf{En $\rightarrow$ Ru}} & \multicolumn{2}{c}{\textbf{En $\rightarrow$ Zh}} & \multicolumn{2}{c}{\textbf{Average}}\\
\cmidrule(lr){2-3}
\cmidrule(lr){4-5}
\cmidrule(lr){6-7}
\cmidrule(lr){8-9}
\cmidrule(lr){10-11}
\cmidrule(lr){12-13}

& Aya  & Llama  & Aya  & Llama  & Aya  & Llama  & Aya  & Llama  & Aya  & Llama & Aya  & Llama \\
\midrule

Random Continuous &0.50&0.50&0.50&0.50&0.50&0.50&0.50&0.50&0.49&0.50&0.50&0.50 \\
\midrule
Probability&0.63&0.63&0.58&0.58&0.60&0.53&0.60&0.62&0.61&0.59&0.60&0.59 \\
Entropy&\bl{0.64}&0.63&\bl{0.59}&0.59&\bl{0.62}&0.53&\bl{0.61}&0.62&\bl{0.62}&0.59&0.62&0.59 \\
\midrule
Word\_Numeric&0.61&\bl{0.65}&0.58&\bl{0.60}&0.60&0.60&\bl{0.61}&\bl{0.63}&0.58&0.65&0.60&0.62 \\
Word\_Likert&0.54&0.60&0.54&0.58&0.52&0.56&0.53&0.59&0.51&0.60&0.53&0.59 \\
Token\_Numeric&0.62&0.63&0.54&0.56&\bl{0.62}&\bl{0.62}&\bl{0.61}&0.62&0.61&\bl{0.68}&0.60&0.62 \\
Token\_Likert&0.57&0.55&0.53&0.53&0.52&0.56&0.58&0.54&0.57&0.58&0.55&0.55 \\

\bottomrule[0.5pt]
\end{tabular}
\caption{AUROC for the continuous methods in their alignment with ground truth on the test set (higher is better). Best performing results are bolded in each column.}
\label{tab:auroc_all_test_results}
\end{table*}

\begin{table*}
\footnotesize
\centering
\begin{tabular}{l|cc|cc|cc|cc|cc|cc}
\toprule[0.5pt]
\multicolumn{1}{c}{\multirow{2}{*}{Method}} & \multicolumn{2}{c}{\textbf{En $\rightarrow$ Cs}} & \multicolumn{2}{c}{\textbf{En $\rightarrow$ Hi}} & \multicolumn{2}{c}{\textbf{En $\rightarrow$ Ja}} & \multicolumn{2}{c}{\textbf{En $\rightarrow$ Ru}} & \multicolumn{2}{c}{\textbf{En $\rightarrow$ Zh}} & \multicolumn{2}{c}{\textbf{Average}}\\
\cmidrule(lr){2-3}
\cmidrule(lr){4-5}
\cmidrule(lr){6-7}
\cmidrule(lr){8-9}
\cmidrule(lr){10-11}
\cmidrule(lr){12-13}
& Aya  & Llama  & Aya  & Llama  & Aya  & Llama  & Aya  & Llama  & Aya  & Llama & Aya  & Llama \\
\midrule
Probability&0.11&0.13&0.06&0.06&0.03&0.05&0.13&0.14&0.06&0.07&0.08&0.09\\
Entropy&\bl{0.13}&\bl{0.14}&\bl{0.07}&0.06&\bl{0.04}&0.04&\bl{0.14}&\bl{0.15}&\bl{0.08}&\bl{0.08}&0.09&0.09\\
\midrule
Word\_Numeric&0.10&0.13&0.06&0.06&0.03&0.05&0.12&0.14&0.05&0.07&0.07&0.09\\
Word\_Likert&0.08&0.13&0.05&\bl{0.07}&0.02&\bl{0.06}&0.09&0.13&0.04&0.07&0.06&0.09\\
Token\_Numeric&0.08&0.11&0.05&0.04&0.03&0.04&0.11&0.13&0.05&0.07&0.06&0.08\\
Token\_Likert&0.08&0.09&0.05&0.04&0.02&0.04&0.11&0.11&0.05&0.05&0.06&0.07\\
\bottomrule[0.5pt]
\end{tabular}
\caption{AUPRC (or Average Precision) for the methods in their alignment with ground truth on the test set (higher is better). Best performing results are bolded in each column.}
\label{tab:average_precision_all_test_results}
\end{table*}

\begin{table*}
\footnotesize
\centering
\begin{tabular}{l|cc|cc|cc|cc|cc|cc}
\toprule[0.5pt]
\multicolumn{1}{c}{\multirow{2}{*}{Method}} & \multicolumn{2}{c}{\textbf{En $\rightarrow$ Cs}} & \multicolumn{2}{c}{\textbf{En $\rightarrow$ Hi}} & \multicolumn{2}{c}{\textbf{En $\rightarrow$ Ja}} & \multicolumn{2}{c}{\textbf{En $\rightarrow$ Ru}} & \multicolumn{2}{c}{\textbf{En $\rightarrow$ Zh}} & \multicolumn{2}{c}{\textbf{Average}}\\
\cmidrule(lr){2-3}
\cmidrule(lr){4-5}
\cmidrule(lr){6-7}
\cmidrule(lr){8-9}
\cmidrule(lr){10-11}
\cmidrule(lr){12-13}

& Aya  & Llama  & Aya  & Llama  & Aya  & Llama  & Aya  & Llama  & Aya  & Llama & Aya  & Llama \\

List with Probability&0.52 	&0.48 	&0.47 	&0.52 	&0.43 	&0.52 	&0.42 	&0.46 	&0.43 	&0.5 	&0.45 	&0.50
\\
\midrule
List with Entropy &0.52 	&0.48 	&0.47 	&0.52 	&0.45 	&0.53 	&0.42 	&0.46 	&0.43 	&0.5 	&0.46 	&0.50 \\
\end{tabular}
\caption{F1 results for the alignment of list method with token probabilities and with token entropies (higher is better). We binarize according to a threshold obtained via the corresponding development set. }
\label{tab:list_verbalized_internal_f1_test_results}
\end{table*}

\begin{table*}
\footnotesize
\centering
\begin{tabular}{l|cc|cc|cc|cc|cc|cc}
\toprule[0.5pt]
\multicolumn{1}{c}{\multirow{2}{*}{Method}} & \multicolumn{2}{c}{\textbf{En $\rightarrow$ Cs}} & \multicolumn{2}{c}{\textbf{En $\rightarrow$ Hi}} & \multicolumn{2}{c}{\textbf{En $\rightarrow$ Ja}} & \multicolumn{2}{c}{\textbf{En $\rightarrow$ Ru}} & \multicolumn{2}{c}{\textbf{En $\rightarrow$ Zh}} & \multicolumn{2}{c}{\textbf{Average}}\\
\cmidrule(lr){2-3}
\cmidrule(lr){4-5}
\cmidrule(lr){6-7}
\cmidrule(lr){8-9}
\cmidrule(lr){10-11}
\cmidrule(lr){12-13}

& Aya  & Llama  & Aya  & Llama  & Aya  & Llama  & Aya  & Llama  & Aya  & Llama & Aya  & Llama \\

List with Probability&0.53&0.59&0.51&0.54&0.54&0.56&0.53&0.59&0.52&0.56&0.53&0.57
\\
\midrule
List with Entropy &0.53&0.59&0.51&0.54&0.54&0.57&0.54&0.59&0.53&0.56&0.53&0.57 \\
\end{tabular}
\caption{AUROC results for the alignment of list method with token probabilities and with token entropies (higher is better). }
\label{tab:list_verbalized_internal_auroc_test_results}
\end{table*}

\begin{figure*}[t]
    \centering
    \begin{subfigure}{0.48\linewidth}
        \centering
        \includegraphics[width=\linewidth]{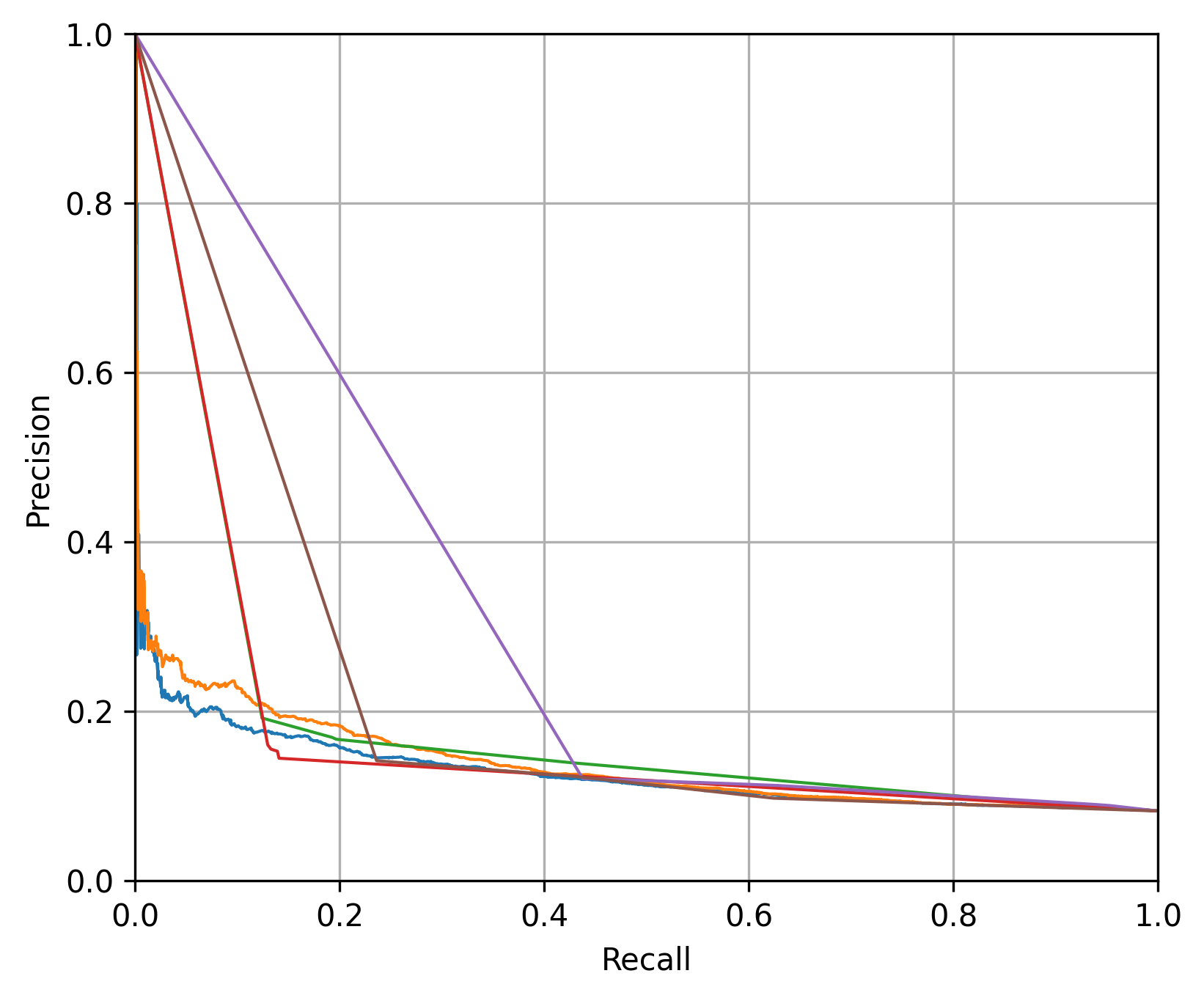}
        \caption{Aya23}
        \label{fig:method_a}
    \end{subfigure}\hfill
    \begin{subfigure}{0.48\linewidth}
        \centering
        \includegraphics[width=\linewidth]{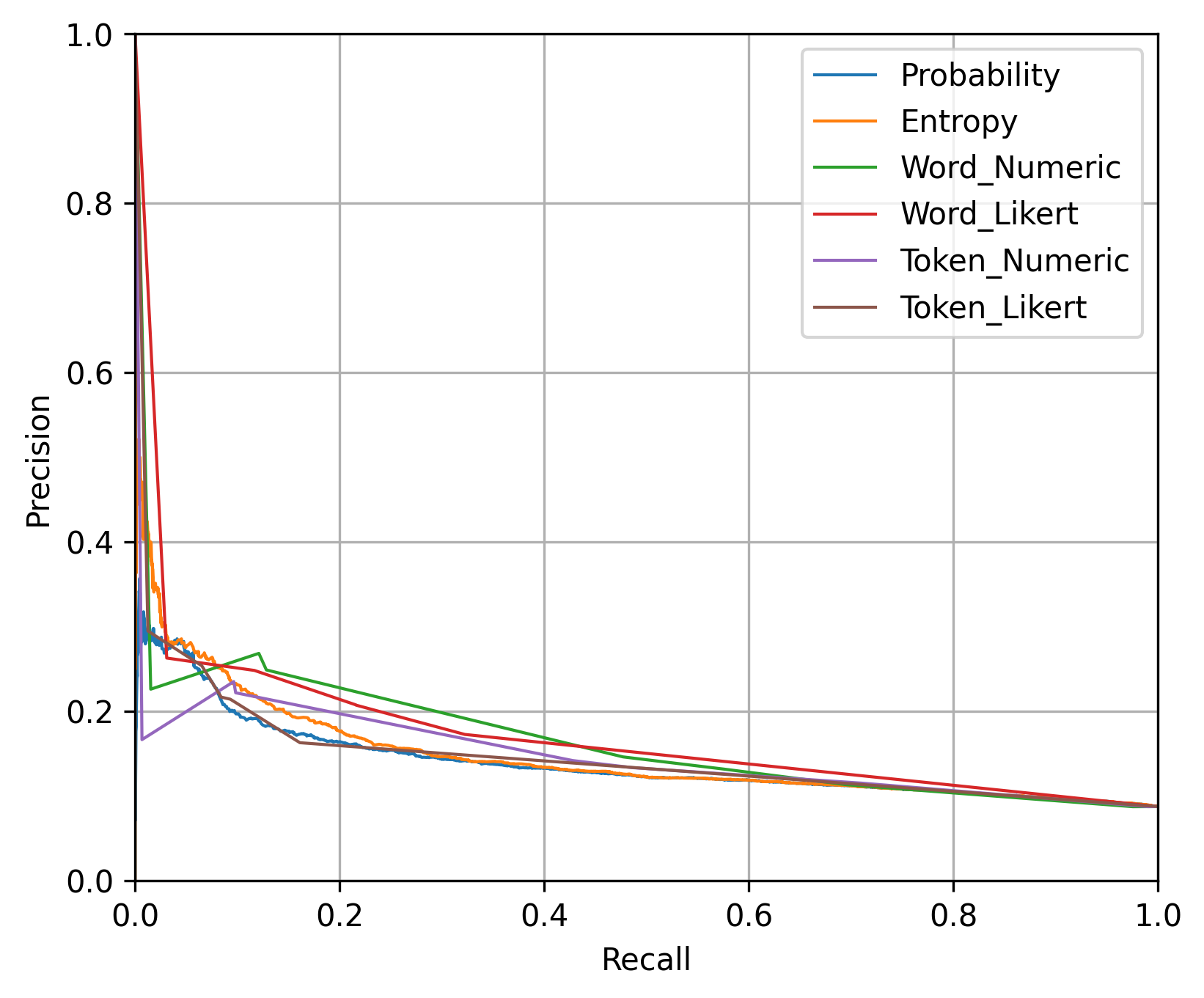}
        \caption{Llama3-70B}
        \label{fig:method_b}
    \end{subfigure}
    \caption{Precision and recall curves of different methods for English $\rightarrow$ Russian translations. We observe that even numeric variants of verbalized methods provide limited granularity, effectively partitioning predictions into a few confidence levels (see Table \ref{tab:numerical_ratios}). }
    \label{fig:side_by_side_pre_rec_cur_russian}
\end{figure*}

\begin{figure*}[h]
    \centering
    \includegraphics[width=1\linewidth]{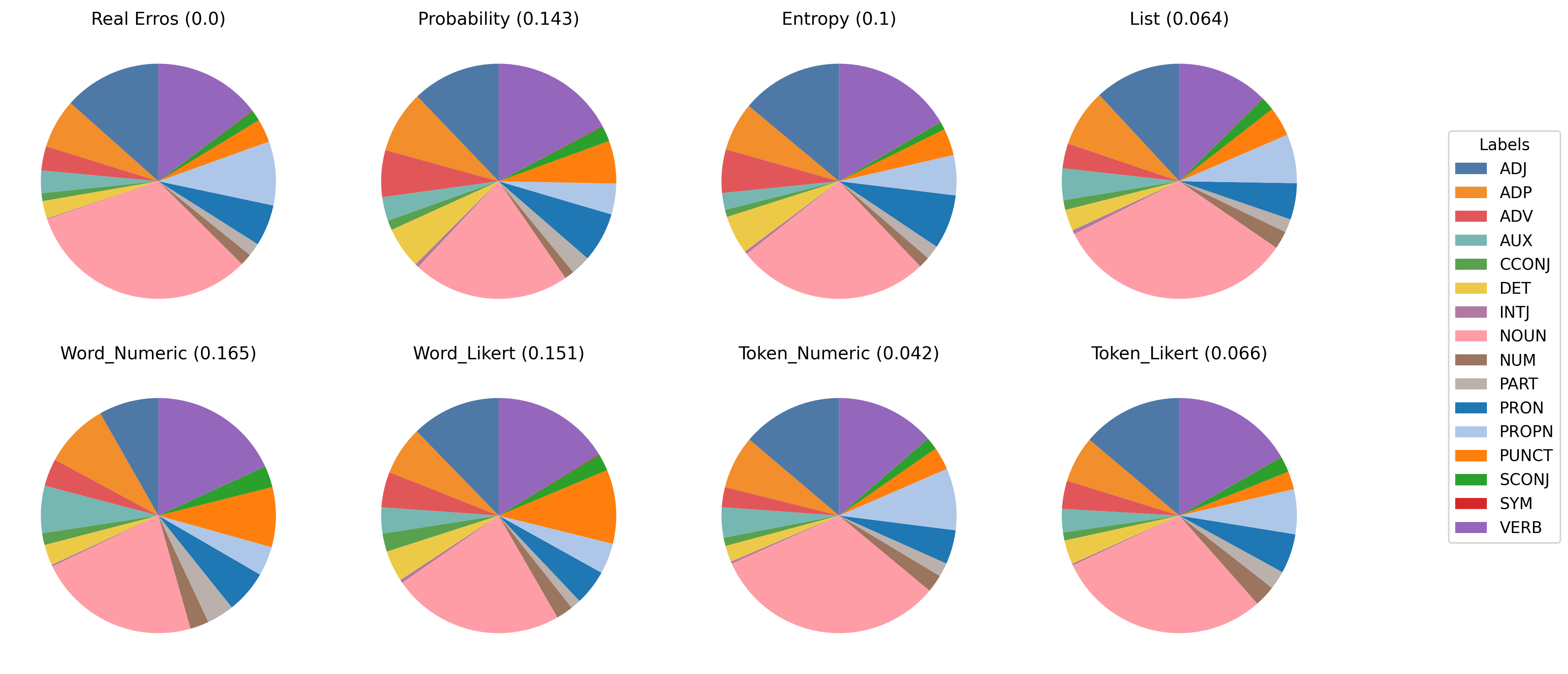}
    \caption{Distributions of different POS tags in the tokens considered as \textit{incorrect} for each method, averaged across all languages on the test set for Aya23. The number in paranthesis is the Hellinger distance of the distribution of the errors as identified by the method from those of the real distribution. }
    \label{fig:aya_pos_piechart1}
\end{figure*}

\begin{figure*}[h]
    \centering
    \includegraphics[width=1\linewidth]{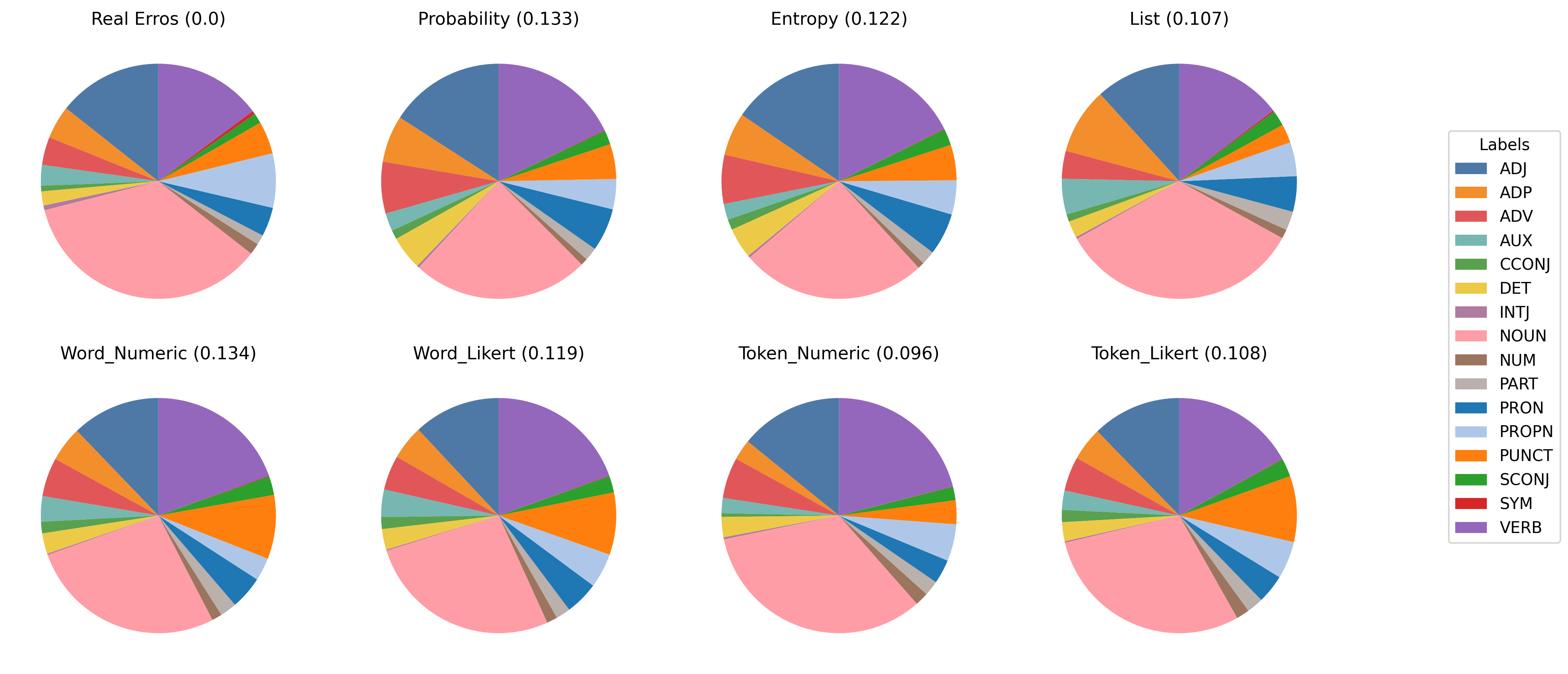}
    \caption{Distributions of different POS tags in the tokens considered as wrong for each method, averaged across languages. The number in paranthesis is the Hellinger distance of the distribution of the errors for the method from those of the real distribution. These are for the test set on Llama3-70B.}
    \label{fig:llama_pos_piechart}
\end{figure*}


\end{document}